\documentclass{kluwer} 
\usepackage{listings}
\usepackage[dvips]{epsfig,graphicx}
\usepackage[dvips]{color}

\newcommand{\mybox}{\mbox{\Large$\Box$}}
\newcommand{\myboxtimes}{\boxtimes}
\newcommand{\myboxdot}{\boxdot}

\newcommand{\mybbox}{\bbox}
\def\bbox{\vrule width 5 true pt height 5 true pt depth 0 pt}
\newtheorem{Theorem}{Theorem}
\newtheorem{Definition}{Definition}
\newtheorem{Proposition}{Proposition}
 \newtheorem{Example}{Example}
\newtheorem{Observation}{Observation}

\newtheorem{Lemma}{Lemma}
\newenvironment{ContinueExample}{{\noindent {EXAMPLE 1} ({\it continued})}\begin{itshape}}{\end{itshape}}

\newcommand{\Do}{{\mbox{\sc Do}}}
\newcommand{\Poss}{{\mbox{\sc Poss}}}
\newcommand{\Holds}{{\mbox{\sc Holds}}}
\newcommand{\Wins}{{\mbox{\sc Wins}}}
\newcommand{\Terminal}{{\mbox{\sc Terminal}}}
\newcommand{\Strat}{{\mbox{\sc Strat}}}
\newcommand{\StratPlayout}{{\mbox{\sc Strategic}}}
\newcommand{\Turn}{{\mbox{\sc Turn}}}
\newcommand{\Winnable}{{\mbox{\sc Win}}}

\renewcommand{\to}{\supset}

\newcommand{\thoughtful}{{\mbox{\rm\em thoughtful\/}}}
\newcommand{\combined}{{\mbox{\rm\em combined\/}}}
\newcommand{\defence}{{\mbox{\rm\em defence\/}}}

\newcommand{\passivedefence}{{\mbox{\rm\em passive\_defence\/}}}
\newcommand{\fillonext}{{\mbox{\rm\em fill\_o\_next\/}}}
\newcommand{\cautious}{{\mbox{\rm\em cautious\/}}}
\newcommand{\fillisolated}{{\mbox{\rm\em fill\_isolated\/}}}
\newcommand{\fillany}{{\mbox{\rm\em fill\_any\/}}}
\newcommand{\fillleftmost}{{\mbox{\rm\em fill\_leftmost\/}}}

\newcommand{\sitc}[3]{{#1}^{SC}[{#2},{#3}]}
\newcommand{\sitcstr}[3]{{#1}[{#2},{#3}]}

 \lstdefinelanguage{ASP}{
  morekeywords={holds, legal, terminal, does, wins},
  sensitive,
  morecomment=[s]{/*}{*/},
}

\lstdefinestyle{ASP}{
  language=ASP,
  numbers=left,
  stepnumber=1,
  firstnumber=last,
  showlines=true,
  numbersep=4pt,
  numberstyle=\tiny,
  tabsize=2, breaklines=false, escapeinside={/*@}{@*/},
  keywordstyle=\bf\ttfamily,
  basicstyle=\scriptsize\ttfamily
}
\long\def\comment#1{}
\newenvironment{proof}{{\bf Proof\/:\ \ \ }}{\ \hfill$\mybbox$}

\newcommand{\Proof}[1]{\begin{proof}{#1}\end{proof}}

\newcommand{\stg}{(N, W, \mathcal{A}, \bar{w}, t, l, u,g)}
\newcommand{\path}{w_0 \stackrel{a_0}{\rightarrow}w_1\stackrel{a_1}{\rightarrow} \cdots\stackrel{a_{m-1}}{\rightarrow}w_m}
\newcommand{\pathe}{w_0 \stackrel{a_0}{\rightarrow}w_1\stackrel{a_1}{\rightarrow} \cdots\stackrel{a_{e-1}}{\rightarrow}w_e}

\begin{document}
\begin{article}
\begin{opening}  
\title{Representing and Reasoning about Game Strategies}
\author{Dongmo \surname{Zhang} }
\institute{The University of Western Sydney, Australia}
\author{Michael \surname{Thielscher}}
\institute{The University of New South Wales, Australia}

%%%%%%%%%%%%%%%%%%%%%%%%%%%%%%%%%%%%%%%%%%%%%%%%%%%%%%%%%%%%%%%%%%%%%%%%%%%%%%%%
%%

%\maketitle

\begin{abstract}
As a contribution to the challenge of building game-playing AI systems, we develop and analyse a formal language for representing and reasoning about strategies. Our logical language builds on the existing general Game Description Language (GDL) and extends it by a standard modality for linear time along with two dual connectives to express preferences when combining strategies. The semantics of the language is provided by a standard state-transition model. As such, problems that require reasoning about games can be solved by the standard methods for reasoning about actions and change. We also endow the language with a specific semantics by which strategy formulas are understood as move recommendations for a player. To illustrate how our formalism supports automated reasoning about strategies, we demonstrate two example methods of implementation\/: first, we formalise the semantic interpretation of our language in conjunction with game rules and strategy rules in the Situation Calculus; second, we show how  the reasoning problem can be solved with Answer Set Programming.
\end{abstract}
\end{opening} 

%%%%%%%%%%%%%%%%%%%%%%%%%%%%%%%%%%%%%%%%%%%%%%%%%%%%%%%%%%%%%%%%
\section{Introduction}

Strategic reasoning has been a major research theme in game theory. However, ``much of game theory is about the question whether strategic equilibria exist'', as Johan van Benthem points out, ``{\it but there are hardly any explicit languages for defining, comparing, or combining strategies'' \comment{ as such the way we have languages for actions and plans, maybe the closest intuitive analogue to strategies.}}\cite{benthem2008praise}. The intrinsic difficulty of modelling strategic reasoning is that reasoning about strategies is not purely deductive but combines temporal reasoning, counterfactual reasoning, reasoning about actions and preferences, and multi-agent interaction. If any logic is used, such a logic must be able to support these reasoning mechanisms.

In recent years a number of specific logical formalisms have been proposed for specifying strategic behaviour of agents in multi-agent systems~\cite{Pauly03,vanderHoek05,ramanujam:dynamic,Chatterjee2010677,MogaveroMV10}. These frameworks offer \comment{ expressive formal representation and} modelling facilities and inference mechanisms for specifying strategy reasoning in multi-agent systems. At the same time, the conceptually simple, general {\it Game Description Language\/} (GDL) has been developed as a practical language for encoding the rules of arbitrary games so that they can be understood by general game-playing systems, whose task is to learn to play unknown games without human intervention \cite{genese:genera,B:66}.

In this paper, we explore the middle ground between pure game specification languages like GDL on the one hand, and existing expressive formalisms for strategic reasoning on the other hand. Our main contributions can be summarised as follows\/:
\begin{enumerate}
  \item We show how a simple extension of GDL using a standard modality for linear time suffices to describe strategies in addition to the mere rules of a game, and we present a specific semantics by which formulas in this language can be understood as move recommendations for a player.
  \item We enrich our language by two {\it preference\/} operators, respectively called prioritised disjunction and prioritised conjunction, and show how to use them to describe strategies that are complete (i.e., provide a move recommendation in every state) and deterministic (i.e., move recommendations are always unique).
  \item We demonstrate two example methods of implementation\/: a formalisation of the semantic interpretation of our language in the Situation Calculus \cite{reiter2001knowledge} and a translation of a subset of our language into Answer Set Programming \cite{gelfon:answer}.
\end{enumerate}
These results are accompanied by a thorough mathematical analysis of the language and its semantics, in particular with regard to our novel preference operators.
The advantage of our framework is to allow for concise but complete representations of games and strategic behaviour of agents so that the approach promises to be practically useful for the design of game-playing agents. We will use a simple game scenario as a running example to explain the basic concepts and to demonstrate how our language can be used to write strategies for game-playing agents. Our results lay the foundations for a variety of applications of automated reasoning about strategies in AI systems, including the following\/:
\begin{itemize}
  \item A general game player can be fed not only with the mere rules of a new game but also with declarative descriptions of tailor-made strategies.
  \item General game players can use a high-level, declarative representation of their strategies as the basis for learning, maintaining, and reasoning about them.
  \item Players can also use the language and its inference mechanism to represent, revise and reason with their beliefs about their opponents' strategies.
\end{itemize}

The remainder of the paper proceeds as follows. In Section~\ref{se:STG} we will begin our technical exposition with the definition of a simple state transition semantics for games. In Section~\ref{s:strategies}, we will then define the semantic concept of a strategy within this framework, and in Section~\ref{s:representation} we will develop a formal game specification language based on GDL for describing both game rules and strategies in a concise manner. In Section~\ref{s:combining}, we will introduce the two preference connectives and show how to use these to combine strategies into more complex ones. In Section~\ref{se:reasoning}, we demonstrate how to verify if a strategy could bring out an expected outcome with the model-based approach.  In Section~\ref{s:computing}, we will encode the semantic interpretation of our language in conjunction with game rules and strategy rules in the Situation Calculus and also illustrate how the reasoning problem can be solved with Answer Set Programming. We conclude with a discussion of related work.
 
%%%%%%%%%%%%%%%%%%%%%%%%%%%%%%%%%%%%%%%%%%%%%%%%%%%%%%%%%%%%%%%%
\section{State Transition Games}
\label{se:STG}
\comment{In game theory, a strategy is defined as a decision of actions under specific histories. Our concept of strategies is more generic. We start with a definition that does not relay on histories.} 

%\subsection{State transition game}
To set the stage for our work, we first define formally a general state transition model for games.
\begin{Definition}\label{de:statetransitiongame}
A \emph{state transition game} $G$ is a tuple $(N, W, \mathcal{A}, \bar{w},$ $ t, l, u,g)$, where
%In a \emph{state transition game} $(N, W, \mathcal{A}, \bar{w}, t, l, u,g)$,
\begin{enumerate}
\item $N$ is a non-empty finite set of players;

\item $W$ is a non-empty set of states (possible worlds);

\item $\mathcal{A}=\bigcup_{i\in N} A^i$, where $A^i$ is a non-empty finite set of \emph{actions} for player \mbox{$i \in N$};

\item $\bar{w}\in W$, representing the \emph{initial} state;

\item $l\subseteq W \times \mathcal{A}$ is a binary \emph{legality} relation, describing what actions are allowed in which states;
 
\item $u:\mathcal{A}\times W\mapsto W$ is an \emph{update} function, specifying the state transitions;
 
\item $g: N\mapsto 2^W$ is a \emph{goal} function, specifying the winning states of each agent;

\item $t \subseteq W$, representing the \emph{terminal} states.
\end{enumerate}
\end{Definition}

To keep our formalism as simple as possible, we assume that all actions are performed asynchronously, each by a single player (although games need not be turn-taking).
%No concurrent actions or simultaneous actions are considered in the current formalism.
We also assume that different agents have different actions, that is, $A^i\cap A^j=\emptyset\,$ for any $i\not=j$, but of course two actions may have the same effects.

A sequence $\path$ is called a \emph{complete path} %in a state transition game $G$
if 
\begin{enumerate}
\item $w_0=\bar{w}$, $w_m\in t$, and $w_j\in W$ for all $0<j< m$;
\item $a_j\in \mathcal{A}$ for all $0\leq j< m$;
\item $(w_j,a_j)\in l$ for all $0\leq j< m$; and
\item $u(a_j,w_j)=w_{j+1}$ for all $0\leq j< m$.
\end{enumerate}

Any segment $w_{k} \stackrel{a_{k}}{\rightarrow}w_{k+1}\stackrel{a_{k+1}}{\rightarrow}\cdots\stackrel{a_{l-1}}{\rightarrow}w_{l}$, where $k\leq l$,  of a complete path is called a \emph{reachable path}. We let $\mathcal{P}(G)$ denote all reachable paths in a state transition game~$G$. Note that a single state without action can be a reachable path (i.e., where $k=l$). We call such a singleton path $w\in W\cap\mathcal{P}(G)$ a {\it reachable state}.

A state-action pair $(w,a)$ is called a {\it reachable legal move} (or simply {\it a move}) in $G$ if there is a $w'$ such that $w\stackrel{a}{\rightarrow}w'$ is a reachable path. The set of all such moves in $G$ is denoted by $\Omega(G)$. More formally, 
\begin{equation}\label{eq:move}
\Omega(G)=\{(w,a): \mbox{there is }w'\in W \mbox{ such that } w\stackrel{a}{\rightarrow}w'\in\mathcal{P}(G) \}  
\end{equation}
Furthermore, the moves for player $i$ are denoted by $\Omega^i(G)$:
\begin{equation}\label{eq:moveplayer}
\Omega^i(G)=\{(w,a)\in \Omega(G): a\in A^i\}
\end{equation}
For convenience, we let
\begin{equation}\label{eq:lw}
l^i(w)=\{a: (w,a)\in \Omega^i(G)\}
\end{equation}
\medskip

To facilitate the presentation of our framework, we will use as our running example a simple two-player game, which we call {\it CrossDot} game\footnote{The game can be viewed as a variation of tic-tac-toe or a simplified Gomoku game.}. 
 \begin{Example}{\bf (CrossDot Game)\ }\label{ex:boxdot1}\rm 
Two players take turns in placing either a cross ``$\times$'' (player~1) or a dot ``$\cdot$'' (player~2) into an empty box in a line of $m$, where $m\geq 2$:
\[
\begin{array}{c}
\underbrace{\,\mybox \mybox \mybox\ \cdots\ \mybox\,} \\
m\\
\end{array}
\]
\noindent
Each box can contain at most one object. The first player to successfully fill $k$ ($1<k\le m$) consecutive boxes will end and win the game, where $k$ is arbitrary but fixed. If all boxes have been filled without a winner, the game ends with a tie. We assume that the ``$\times$''-player goes first.

To describe the scenario in terms of the state transition game, let $N=\{1,2\}$ be the players and
\[
 W\!=\!\{ (t_1,t_2, x_1,x_2,\ldots,x_m)\!: t_1,t_2\!\in\! \{0,1\}\ \&\  x_1,\ldots,x_m\!\in\! \{\mybox,\myboxtimes,\myboxdot\} \}
 \]
the set of possible states, where $t_1,t_2$ specify whose turn it is ($t_i=1$ if it is player $i$'s turn; otherwise $t_i=0$) and
%\footnote{Note that such a representation is also applicable to the case when simultaneous moves are allowed.}.  
$x_1,\ldots,x_m$ indicate the status of the boxes. The initial state is $\bar{w}=(1,0, \mybox,\mybox, \cdots,\mybox)$.
% because we assume the first player takes the first move.
 
We write $a^i_j$ to denote the action of player $i$ marking the $j$th box. Let $A^i=\{a^i_j: i\in N\ \&\  1\leq j\leq m\}$. 
%The set of all the actions for both players is denoted by $A$. {\bf [MICHAEL: Should this be $\mathcal{A}$? Do we need that here?]} 
We refrain from explicitly listing the legality relation, the update function, and the terminal and goal states for the players as this is possible but considerably lengthy even for a very simple game like this; the syntactic \emph{axiomatision} of this game given in the following section will be much more concise and practical. Let us just pick some random examples: For \mbox{$m=4$} and $k=2$, the state $(1,0,\myboxdot, \myboxtimes, \myboxtimes, \myboxdot)$ is a terminal state and also a goal state for player 1. Under our assumption that player~1 takes the first move, the state-action pair $((1,0,\myboxdot, \mybox, \myboxtimes, \mybox), a^1_2)$, say, satisfies the legality relation, whereas $((1,0,\myboxdot, \mybox, \myboxtimes, \mybox), a^2_2)$ does not because it is not player~2's turn at this stage. As an example of the update function we have $u(a^1_2, (1,0,\myboxdot, \mybox, \myboxtimes, \mybox)) = (0,1,\myboxdot, \myboxtimes, \myboxtimes, \mybox)$.

For future reference, by $G_{\mbox{\scriptsize\rm\em CrossDot}}^{m,k}$ we denote the instance of our game that consists of $m$ boxes and has winning length $k$.
\hfill $\mybbox$
\end{Example}

%%%%%%%%%%%%%%%%%%%%%%%%%%%%%%%%%%%%%%%%%%%%%%%%%%%%%%%%%%%%%%%%
\section{Strategies in State Transition Games} \label{s:strategies}

%In game theory, a player's strategy in a game determines the action the player will take at any state of the game. 
\comment{
 a player's strategy in a game is a complete plan of action for whatever situation might arise
A strategy defines a set of moves or actions a player will follow in a given game. A strategy must be complete, defining an action in every contingency, including those that may not be attainable in equilibrium. Such moves may be random, in the case of mixed strategies.
}
In a multi-agent game environment, agents strive to achieve their goals. How individual agents act is determined by their \emph{strategies}: At a  state of a game, the strategies that the agent applies determine which actions the agent will take.  For example, a strategy of a player in the game of chess reflects which moves the player would play in certain positions that can be reached according to the standard chess rules.  In terms of the state transition game, a strategy of a player can be defined as a relation between game states and (legal) actions the player will perform.  \comment{A strategy of an agent is a plan of action for any given world state. In the following we provide a formal definition of strategies in semantic terms. Prior to this we introduce our first running example.}
\begin{Definition}\label{de:strategy}
A \emph{strategy} $S$ of player $i$ is a subset of $W\times A^i$ such that $S\subseteq \Omega^i(G)$. 
%In other words, a strategy of a player is a collection of their researchable steps. 
%Each $(w,a)\in S_i$ is called a \emph{strategy} of player $i$ under the strategy rule $S_i$. 
\end{Definition}
Intuitively, a strategy of a player specifies which actions the player should take in which states. Note that our concept of strategy is significantly different from the ones in the context of alternating-time temporal logic (ATL)~\cite{AlurHK02,vanderHoek05,Waltheretal07}. In ATL, a strategy is a function that maps each state (or a sequence of states) to an action. In other words, a strategy specifies which action has to do exactly in each state or each history of states. However, our concept of strategy can express a ``rough idea'' of what to do.  A strategy may suggest no action, one action or several actions to do in each state. A player might apply several strategies in one game. A single strategy does not necessarily determine the moves for all possible legal positions of a game. In the later sections, we will further develop a technology to combine and refine strategies in order to generate a strategy with desirable properties.

\subsection{Properties of Strategies}
\label{subse:propertiesOfStrategies}
We say that a strategy $S$ is \emph{valid} if $S\neq \emptyset$. A strategy $S$ of player~$i$ is \emph{complete} if for each reachable state $w\in W\cap\mathcal{P}(G)$, there is a move $(w,a^i)\in S$ unless $l^i(w)=\emptyset$. In other words, a complete strategy provides the player with a ``complete'' guideline that always provides the player with one or more suggestions how to act when it is his move.
A strategy $S$ is \emph{deterministic} if for any $(w,a)\in S$ and $(w,a')\in S$, we have $a=a'$.
%In other words, a deterministic strategy defines at most one action for each state. 
A strategy is \emph{functional} if it is complete and deterministic.
An agent with a functional strategy knows precisely what to do in any reachable game state.  For instance, if an agent~$i\in N$ has a default action $a^i_0\in A^i$ that is always legal in any reachable state, then the following simple strategy for this player is complete and deterministic, hence functional:
$$S^i=\{(w,a^i_0): w\in W\cap \mathcal{P}(G)\}$$
\comment{
Note that in real game playing, defining a functional strategy for a player is not always easy because different situations (states) may require different actions to apply. \\
}
\comment{
The following concept will be useful in the sequence. 
\begin{Definition}\label{de:complywith}
Given a game $G$ and a strategy $S$ of player $i$, we say that a reachable path $\path$ {\it complies with strategy $S$ for player $i$} if $(w_k, a_k)\in S$ for all $k$ ($0\leq k<m$) such that $a_k\in A^i$.\footnote{Specifically, the path complies with any $S$ if $m=0$.}
\end{Definition}
}
\begin{Example}\label{ex:boxdot2}\rm
Consider an instance, $G_{\mbox{\scriptsize\rm\em CrossDot}}^{4,2}$,  of the CrossDot  game.  The following is an example of strategies for player~1 that intuitively says {\it ``fill a box next to one you have marked before''}\/:
$$ \begin{array}{lll}
S^1=\{\!\!\! & ((1,0,\myboxtimes, \mybox, \myboxdot, \mybox), a^1_2), & \!((1,0,\myboxtimes, \mybox, \mybox,\myboxdot), a^1_2), \\
&  ((1,0,\myboxdot, \myboxtimes, \mybox, \mybox), a^1_3), & \! ((1,0,\mybox, \myboxtimes,\myboxdot,\mybox), a^1_1),  \\
& ((1,0, \mybox, \myboxtimes, \mybox, \myboxdot), a^1_1), & \! ((1,0,\mybox, \myboxtimes,\mybox,\myboxdot), a^1_3), \\
& ((1,0,\myboxdot, \mybox, \myboxtimes, \mybox), a^1_2),  & \! ((1,0,\myboxdot, \mybox, \myboxtimes, \mybox), a^1_4), \\
& ((1,0,\mybox, \myboxdot, \myboxtimes, \mybox), a^1_4), & \! ((1,0,\mybox, \mybox, \myboxtimes, \myboxdot), a^1_2),\\
& ((1,0,\myboxdot, \mybox, \mybox, \myboxtimes), a^1_3),   & \!((1,0,\mybox, \myboxdot, \mybox, \myboxtimes), a^1_3)\,\}
\end{array} $$
It is easy to see that the strategy is valid but neither complete nor deterministic.
\hfill $\mybbox$
\end{Example}

%%%%%%%%%%%%%%%%%%%%%%%%%%%%%%%%%%%%%%%%%%%%%%%%%%%%%%%%%%%%%%%%
\section{Strategy Representation and Semantics} \label{s:representation}

The above example shows that directly representing a strategy in a state transition game requires to list every move that complies with the strategy. In the following, we develop a syntactical representation that allows to describe a strategy much more concisely.

\subsection{Formal Game Specification Language}

We first present a logic-based, general game description language with linear time.
\begin{Definition}\rm
Consider a propositional modal language $\mathcal{L}$ with these components:
\begin{itemize}
\item a non-empty finite set $\Phi$ of propositional variables;

\item a non-empty finite set $N$ of agent symbols;

\item a non-empty finite set $A^i$ of action symbols for each $i\in N$;

\item propositional connectives $\neg$, $\wedge$, $\vee$, $\rightarrow$ and $\equiv$;\footnote{Only $\neg$ and $\wedge$ are treated as primitives.%
%The other two can be defined using truth value table
}

%\item  pseudo-function symbol $true(.)$;

\item pseudo-function symbols $does(.)$, $legal(.)$, and $wins(.)$;

\item modal operator $\bigcirc$;

\item special propositional symbols $init$ and $terminal$.
\end{itemize}
\end{Definition}
Formulas in $\mathcal{L}$ are defined as follows:
\[ \varphi\ :=\ p \ |\  \neg \varphi \ |\  \varphi \wedge \varphi\ |\  \bigcirc \varphi \ |\  does(a)\ |\ legal(a) \ |\ wins(i) \ |\ init\ |\ terminal \]
where $p\in \Phi$, $i\in N$ and $a\in \mathcal{A}=\bigcup_{i\in N} A^i$.\\

Note that we overload $N$ and $A^i$ as they occur in both syntax and semantics. They can be distinguished from the context. As in Definition~\ref{de:statetransitiongame}, we assume that $A_i\cap A_j=\emptyset$ if $i\not=j$. 

We call $does$, $legal$ and $wins$ pseudo-functions because formally each instance $does(a)$, $legal(a)$ or $wins(i)$ is taken to be an individual propositional symbol.
This language is a direct adaptation of the general {\em Game Description Language\/}~\cite{genese:genera} and allows to describe games in a compact way.\\

\begin{ContinueExample}\footnote{While all examples in this paper are based on the {\it CrossDot} game, some examples are specifically numbered just for the purpose of cross-referencing.}\label{ex:crossdot3}\rm\ 
%Consider the game $G^{m,k}_{\mbox{\scriptsize\rm\em CrossDot}}$ in Example~\ref{ex:boxdot1}.
\comment{We overload $N$ to represent the set of agent symbols and $A^i$ the set of action symbols for each player ($i\in N$).}
We use the propositional symbols $p^i_j$ to represent the fact that box $j$ is filled with $i$'s marker, where $i\in \{1,2\}$ and $j\in \{1,\ldots, m\}$. In addition, we use two specific propositional symbols $turn(1)$ and $turn(2)$ to represent players' turns, respectively. Putting all the propositional symbols together, we have $\Phi_{\mbox{\scriptsize\rm\em CrossDot}}=\{p^i_j: 1\leq i\leq 2\ \&\ 1\leq j\leq m\}\cup \{turn(i): i=1 \mbox{ or } 2\}$. With this, we are able to describe the game rules in our logical language.

To begin with, the following rules specify the initial game state:
\begin{eqnarray}
 init\! & \rightarrow & \!\neg p^i_j \ \ \ \ \ \ \ \ \ \ \ \ \ \ \ \ \ \ \ \ \ \ \ \ \ \ \ \ \mbox{ for all } i\in N \mbox{ and } j\leq m \label{eq:2} \\
 init\! & \rightarrow & \!turn(1)\wedge \neg turn(2)
\end{eqnarray}
The following statement defines the winning conditions: {\it Player $i$ wins if there is $j$ such that $j+k-1\le m$ and $p^i_j\wedge \cdots\wedge p^i_{j+k-1}$,}  i.e.  
\begin{equation} \label{eq:3}
wins(i)\,\equiv\,\textstyle\bigvee\limits_{j=1}^{m-k+1} \textstyle\bigwedge\limits_{l=j}^{j+k-1} p^i_l
\end{equation}
With this, the condition for termination is:
\begin{equation}\label{eq:4}
terminal\,\equiv\, wins(1)\vee wins(2)\vee \textstyle\bigwedge\limits_{j=1}^{m}(p^1_j\vee p^2_j)
\end{equation}
As before, let $a^i_j$ denote the action of player $i$ filling box $j$.  
\comment{Furthermore, let $a^1_{nop}$ and $a^1_{nop}$ denote the no-operation action for each player respectively.}
Legality of the actions of each player $i$ can be described thus:
\begin{eqnarray}
& \neg (p^1_j\vee p^2_j)\wedge turn(i)\wedge \neg terminal\,\equiv\, legal(a^i_j) \label{eq:8}
\end{eqnarray}
The effects of the actions are given by\footnote{To avoid too much complexity, we follow~\cite{genese:genera} not offering any solution to the frame problem here. Game descriptions simply include all necessary frame axioms.}
\begin{eqnarray}
& p^i_j\vee does(a^i_j)\equiv \bigcirc p^i_j \label{eq:10} \\
& turn(1) \rightarrow \bigcirc\neg turn(1)\wedge \bigcirc turn(2) \label{eq:11} \\
& turn(2) \rightarrow\bigcirc\neg turn(2)\wedge \bigcirc turn(1) \label{eq:12}
\end{eqnarray}
Let $\Sigma^{m,k}_{\mbox{\scriptsize\rm\em CrossDot}}$ be the set of axioms (\ref{eq:2})--(\ref{eq:12}).
\hfill $\mybbox$\\
\end{ContinueExample}

In the following, we interpret the language based on the state transition model. Give a state transition game $G=\stg$ (see Definition~\ref{de:statetransitiongame}), 
 a {\it valuation function} $v:W\mapsto 2^\Phi$ specifies which atom propositions are true at each state. Propositional formulas and their truth values can be defined accordingly.
\begin{Definition}\label{de:linearmodel}
Let $G=\stg$ be a state transition game and $v$ a valuation function. We call the pair $M=(G,v)$ a \emph{state transition model}. Let $\delta=\path\in \mathcal{P}(G)$ be a reachable path in $G$ and $\varphi$ a formula. We say that
$\delta$ {\em satisfies} $\varphi$ under $M$ (written $M,\delta \models \varphi$) according to the following definition:
\begin{tabbing}
--------\=---------------------------------\=-----\=\kill
\>$M,\delta \models p$                       \>iff\> $p\in v(w_0)$ \quad ($p \in \Phi$)\\
\>$M,\delta \models does(a)$                       \>iff\> $a= a_0$\\
\>$M,\delta \models init$                       \>iff\> $w_0= \bar{w}$\\
\>$M,\delta \models terminal$ \> iff\> $w_0\in t$\\
\>$M,\delta \models legal(a)$ \> iff\> $(w_0,a)\in l$\\
\>$M,\delta \models wins(i)$ \> iff\> $w_0\in g(i)$\\
\>$M,\delta \models \neg \varphi$               \>iff\> $M,\delta \not\models \varphi$\\
\>$M,\delta \models \varphi_1 \wedge \varphi_2$      \>iff\> $M,\delta \models \varphi_1$ and $M,\delta \models \varphi_2$\\
\>$M,\delta \models \bigcirc \varphi$               \>iff\>
$M,w_1 \stackrel{a_1}{\rightarrow}\cdots\stackrel{a_{m-1}}{\rightarrow}w_m\models \varphi$
\end{tabbing}
\end{Definition}

It is worth clarifying that in the limit case $m=0$ (i.e., $\delta=w_0$) we have that\comment{$M, \delta\not\models does(a)$ for any $a\in \mathcal{A}$. However, $M,\delta \models legal(a)$ can be true as long as $(w_0,a)\in l$. } $M,\delta \models does(a)$ and $M, \delta\models \bigcirc\varphi$ hold for any $a$ and $\varphi$. 

A formula $\varphi$ is {\it valid} in model $M$, denoted $M\models \varphi$, if it is satisfied by any reachable path in the game, that is, $M,\delta\models \varphi$ for all $\delta \in \mathcal{P}(G)$. Let $\Sigma$ be a set of sentences in $\mathcal{L}$, then $M$ is a {\it model} of $\Sigma$ if $M\models \varphi$ for all $\varphi\in \Sigma$.\\

\begin{Observation}\label{ob:first}
Consider the CrossDot game $G=G_{\mbox{\scriptsize\rm\em CrossDot}}^{m,k}$ introduced in Example~\ref{ex:boxdot1}.  Let $v$ be a valuation function such that for each state $w=(t_1,t_2,x_1,\cdots,x_m)\in W$, $v(w)=\{turn(i): t_i=1\}\cup \{p^1_j: x_j = \myboxtimes\ \& \ 1\leq j\leq m\}\cup \{p^2_j:x_j=\myboxdot\ \& \ 1\leq j\leq m\}$. Let $M=(G,v)$. Then $M$ is a model of $\Sigma^{m,k}_{\mbox{\scriptsize\rm\em CrossDot}}$ (see equations (\ref{eq:2})--(\ref{eq:12})).
\end{Observation}
\Proof{
Given any reachable path $\delta=\pathe$, we only have to verify that each axiom in $\Sigma^{m,k}_{\mbox{\scriptsize\rm\em CrossDot}}$ is satisfied by $\delta$ in $M$. This is straightforward. Consider for instance the proof that $M,\delta \models init \rightarrow \neg p^i_j$ for any $i\in N$ and $j<m$\/: If $w_0\not =\bar{w}$, this holds trivially because  $M,\delta \models \neg init$. If $w_0 = \bar{w}$, then  $M,\delta \models init$ and by the definition of $v$, $p^i_j\not\in v(w_0)$, hence $M,\delta \models \neg p^i_j$ for any $i\in N$ and $j<m$. The other axioms can be verified similarly.
}\\

In order to develop a syntactical representation for strategies, we introduce the following specific concepts.
Let $\delta=w_0 \stackrel{a_0}{\rightarrow}w_1\stackrel{a_1}{\rightarrow}\cdots\stackrel{a_{m-1}}{\rightarrow}w_m$. We call $\delta$ a \emph{reachable path starting with the move $(w_0,a_0)$}. The set of all the reachable paths in game $G$ that starts with $(w, a)$ is denoted by $(w,a)^\leadsto$. 
Given a state transition model $M=(G,v)$, for any move $(w,a)\in \Omega(G)$, a formula $\varphi$ is \emph{valid under move ($w, a$)}, denoted by $M\models_{(w,a)}\varphi$, if $M,\delta\models \varphi$ for all reachable path $\delta\in (w,a)^\leadsto$.  

%\begin{ContinueExample2}
\begin{Example}\label{ex:boxdot3}\rm\
Consider the same instance of the CrossDot game $G_{\mbox{\scriptsize\rm\em CrossDot}}^{4,2}$ as in Example~\ref{ex:boxdot2}. Let $M$ be the state transition model defined in Observation~\ref{ob:first} for the case $m=4$ and $k=2$.  Assume that $(w,a)=((1,0,\myboxtimes, \myboxdot,\mybox,\mybox),a^1_3)$. It is easy to verify the following:
\begin{tabbing}
--------\=---------------------------------\kill
\> $M\models_{(w,a)}legal(a^1_4)$\\
\> $M\not\models_{(w,a)}does(a^2_3)$\\
\> $M\models_{(w,a)}\neg (p^1_3\vee p^2_3)\wedge \bigcirc p^1_3\wedge\bigcirc (legal(a^2_4)\wedge does(a^2_4))$\\
\> $M\models_{(w,a)}\bigcirc\bigcirc terminal$
\end{tabbing}
\hfill $\mybbox$
%!!!!!!!!!!!!
%How about ``$M\models_{(w,a)} \neg \bigcirc \bigcirc \neg terminal$" ?
%!!!!!!!!!!
\end{Example}
%\end{ContinueExample2}

\comment{
Given a strategy $S$, we write ``$M\models_{(w,a)}\varphi$ {\em complying with~$S\,$\/ for player $i$}'' if $M,\delta\models \varphi$ for all reachable path $\delta\in (w, a)^\leadsto$ that complies with $S$ for player $i$ (see Definition~\ref{de:complywith}).
}

\subsection{Describing Strategies}

We now turn to the syntactical representation of strategies using the language introduced above. 
% Given a state transition model $M$ and any formula $\varphi\in \mathcal{L}$, we 
For any state transition model \mbox{$M=(G,v)$} and formula $\varphi\in\mathcal{L}$, let 
\begin{equation}\label{eq:sofr}
\begin{array}{ll}
S^i(\varphi)=  \{(w,a)\in\Omega^i(G): M\models_{(w,a)} \varphi\}
 \end{array}
\end{equation} 
In other words, $S^i(\varphi)$ comprises all moves for player $i$ under which $\varphi$ is valid.

\begin{Definition}\label{de:strategySemantics}
Given a state transition model $M$, let $S$ be a strategy of player $i$  according to Definition~\ref{de:strategy}. A formula $\varphi$ in $\mathcal{L}$ is a \emph{representation} of $S$ iff $ S= S^i(\varphi)$.
\end{Definition}
In the following, we will call ``strategy'' both a set of moves (i.e., a strategy in the sense of Definition~\ref{de:strategy}) and its representation (using~$\mathcal{L}$). They should be easy to distinguish from the context.  
Note that a formula $\varphi\in \mathcal{L}$ can represent different strategies for different players. 
For instance, the tautology $\top$ can be a strategy of player $i$, i.e., $S^i(\top)$, that allows the player to take any reachable moves at any reachable state. Another example is $does(a^i)$, where $a^i\in A^i$. If representing a strategy of player~$i$, it means to take $a^i$ only at any reachable state. However, if representing a strategy of other players rather than $i$, it means to do nothing. 

%\footnote{This strategy is incomplete if there is a reachable state in which $a^i$ is illegal.}
 
\comment{Since the language we consider in this paper is propositional and finite, the total number of actions and states must be finite, too. It follows that any strategy has a representation in $\mathcal{L}$, which shows that this language is sufficiently expressive for our purpose.}

While the language we consider in this paper is propositional and finite, a state transition game can have infinitely many worlds which cannot be distinguished by the propositions that hold in them. A strategy that assigns different actions to two indistinguishable worlds cannot be described in our language. However, every strategy that is {\em Markovian\/} in the following sense does have a representation in $\mathcal{L}$. 

\begin{Definition}
Given a state transition model $M=(G,v)$, a strategy $S$ is {\em Markovian} if for all $(w_1,a)\in S$ and $w_2\in\mathcal{P}(G)$ such that $v(w_1)=v(w_2)$, $(w_2,a)\in S$.
\end{Definition}

\begin{Proposition}\label{pr:Markovian}
Given a state transition model $M=(G,v)$ of $\mathcal{L}$, any Markovian strategy $S$ of a player has a representation in $\mathcal{L}$. 
\end{Proposition}
\begin{proof} For each $(w,a)\in S$, since the set $\Phi$ of propositional variables is finite, $v(w)$ and $\Phi\setminus v(w)$ are both finite.  Thus the following is a well-formed propositional formula:
  \begin{equation}\label{eq:conjunction}
  \varphi(w,a)=(\bigwedge_{p\in v(w)}p)\wedge(\bigwedge_{p\in \Phi\setminus v(w)}\neg p)\wedge does(a)
  \end{equation}
  
Again $\mathcal{A}$ being finite implies the set $\{\varphi(w,a): (w,a)\in S\}$ to be finite even though $S$ may be infinite. Therefore the following is also a well-formed propositional formula:
$$\varphi = \bigvee\limits_{(w,a)\in S} \varphi(w,a) \comment{= \bigvee\limits_{(w,a)\in S}  [(\bigwedge_{p\in v(w)}p)\wedge(\bigwedge_{p\in \Phi\setminus v(w)}\neg p)\wedge does(a)]}$$
Now we show $\varphi$ is a representation of $S$. Obviously, for any $(w,a)\in S$,  $M\models_{(w,a)} \varphi$. For any $(w,a)\in \Omega^i(G)$, where $i$ is the player under consideration, assume that $M\models_{(w,a)} \varphi$.  By the construction of $\varphi$, there is a move $(w',a')\in S$ such that 
$$M\models_{(w,a)} (\bigwedge_{p\in v(w')}p)\wedge(\bigwedge_{p\in \Phi\setminus v(w')}\neg p)\wedge does(a')$$
It turns out that $v(w)=v(w')$. Since $S$ is Markovian, we have $(w,a')\in S$.  Note that $M\models _{(w,a)} does(a')$ implies $a=a'$ by Definition~\ref{de:linearmodel}.  We then have $(w,a)\in S$, as desired.
\end{proof}\\

Note that Markovian strategies are history-independent. If we want to specify strategies in which move choices depend on the history of a game, then we need to extend our language with syntactic means to talk about past states. This can be done by adding the inverse of the operator $\bigcirc$, written $\bigcirc^{-1}$, to denote that some property holds in ``{\it the previous state}''. We leave this extension for future work.

The reader is reminded that formulas in our language have been endowed with two different semantics. If it is used to represent a property, it has a truth value as normal propositional formula. If a formula is used to represent a strategy, then it no longer has a truth value but represents a set of moves for a player. \comment{Moreover, the same formula represents different strategies for different players. For instance, if we use the constant proposition $\top$ (${=}_{def}\  p\vee \neg p$) to represent a strategy for a player, it means all moves that are available to that player at each state. Obviously different players have different available moves.}\\

 The following observation shows how our language can be used to describe a useful strategy for our running example. 
Compared to its semantical representation (cf. Example~\ref{ex:boxdot2}), the syntactical expression of the strategy is much more compact and meaningful. 
 
\begin{Observation}
Strategy $S^1$ from Example~\ref{ex:boxdot2} is represented by
\[ 
  \varphi=\bigvee_{1< j\leq 4} (p^1_{j-1} \wedge\neg p^1_j\wedge \neg p^2_j \wedge   does(a^1_{j})) 
   \vee\ \bigvee_{1\leq j< 4}(\neg p^1_j\wedge \neg p^2_j \wedge p^1_{j+1}  \wedge does(a^1_{j}))
 \]
%\hfill $\mybbox$
\end{Observation}
\Proof{First we show $S^1\subseteq S^1(\varphi)$. This can be done by verifying  $M\models_{(w,a)}\varphi$ one by one for each $(w,a)\in S^1$.  For instance, consider $w=(1,0,\myboxtimes, \mybox,  \myboxdot,\mybox,)$ and $a=a^1_2$. Then $M\models_{(w,a)} p^1_1\wedge \neg p^1_2 \wedge \neg p^2_2$ and $M\models_{(w,a)} does(a^1_2)$. Thus $M\models_{(w,a)}\varphi$.

In order to show that $S^1(\varphi)\subseteq S^1$, we can prove that for each reachable move $(w,a)$ of player 1, $(w,a)\not\in S^1$ implies $M\not\models_{(w,a)} \varphi$. This can be done by enumerating all reachable moves in $\Omega^1(G^{4,2}_{\mbox{\scriptsize\rm\em CrossDot}})\setminus S^1$. For instance, let $w_0=(1,0,\myboxtimes,\myboxdot, \mybox, \mybox)$ and $a_0=a^1_3$. Obviously, $(w_0,a_0)$ is a reachable move and $(w_0,a_0)\not\in S^1$. To show $M\not\models_{(w_0,a_0)} \varphi$, consider the reachable path $\delta = w_0 \stackrel{a_0}{\rightarrow}w_1$, where $w_1=(0,1,\myboxtimes,\myboxdot, \myboxtimes, \mybox)$. It is easy to verify that $M,\delta\models \neg p^1_2 \wedge \neg p^1_4\wedge does(a^1_3)$ while $M, \delta\models \neg does(a')$ for any $a'\not = a^1_3$. Thus we have $M, \delta\not\models \varphi$. We then yield that $M\not\models_{(w_0,a_0)} \varphi$. Other cases can be verified similarly.
}\\

The above observation shows that we can still use our logical sense to design a strategy despite the significant differences of semantics between a propositional formula and a strategy representation. When doing so it is important to keep it in mind that a propositional formula bears a very different meaning when understood as a strategy rather than a state description.
In the following we demonstrate the speciality of strategy representation with a few examples based on the CrossDot game $G=G_{\mbox{\scriptsize\rm\em CrossDot}}^{4,2}$.
\begin{enumerate}
\item {\it One formula can represent different strategies for different players:}

$\begin{array}{ll}
\!S^1(\bigcirc p^1_1) = & \{(1,0,\mybox, x_2, x_3, x_4), a^1_1)\!\in\! \mathcal{P}(G): x_2,x_3,x_4\! \in\!\{\mybox,\myboxtimes,\myboxdot\}\}\\
& \cup\{(1,0,\myboxtimes, \mybox, x_3, x_4), a^1_2)\in \mathcal{P}(G): x_3,x_4 \in\{\mybox,\myboxtimes,\myboxdot\}\}\\
& \cup\{(1,0,\myboxtimes, x_2, \mybox, x_4), a^1_3)\in \mathcal{P}(G): x_2,x_4 \in\{\mybox,\myboxtimes,\myboxdot\}\}\\
& \cup\{(1,0,\myboxtimes, x_2, x_3, \mybox), a^1_4)\in \mathcal{P}(G): x_2,x_3 \in\{\mybox,\myboxtimes,\myboxdot\}\}
\end{array}
$

$
\begin{array}{ll}
S^2(\bigcirc p^1_1)=\!\! & \{(0,1,\myboxtimes, \mybox, x_3, x_4), a^2_2)\in \mathcal{P}(G): x_3,x_4 \in\{\mybox,\myboxtimes,\myboxdot\}\}\\
& \cup \{(0,1,\myboxtimes, x_2, \mybox, x_4), a^2_3)\in \mathcal{P}(G): x_2,x_4 \in\{\mybox,\myboxtimes,\myboxdot\}\}\\
& \cup\{(0,1,\myboxtimes, x_2, x_3, \mybox), a^2_4)\in \mathcal{P}(G): x_2,x_3 \in\{\mybox,\myboxtimes,\myboxdot\}\}
\end{array}
$

In other words, if $\bigcirc p^1_1$ represents a strategy of player 1, it means that player 1 is to fill the first box with his marker if the box is currently empty or to fill any other empty box if the box has already been filled with a cross whenever it is his turn. \comment{Note that this strategy does not specify any action for the player to do if the first box has been filled with a dot.} However, if the formula $\bigcirc p^1_1$ represents a strategy of player 2, it means player 2 wishes the first box to be filled by player 1. In this case, he waits for the states to come and then do whatever is feasible to him. 

\item {\it Forcing another player into a particular action:}

$
\begin{array}{ll}
&S^1(\bigcirc does(a^2_1))\\ 
&\ \ \ = \{(1,0,\mybox, \mybox, x_3, x_4), a^1_2)\in \mathcal{P}(G): x_3,x_4 \in\{\myboxtimes,\myboxdot\}\}\\
&\ \ \ \ \ \   \cup \{(1,0,\mybox, x_2, \mybox, x_4), a^1_3)\in \mathcal{P}(G): x_2,x_4 \in\{\myboxtimes,\myboxdot\}\}\\
&\ \ \ \ \ \   \cup \{(1,0,\mybox, x_2, x_3, \mybox), a^1_4)\in \mathcal{P}(G): x_2,x_3 \in\{\myboxtimes,\myboxdot\}\}
\end{array}
$

In general, a player does not have control on the other player's actions. But a player may be able to create a situation in which the opponent has no choice other than the desired action. In the above example, player 1 is enforcing a situation in which player 2 has no other option but to perform $a^2_1$ when it becomes his turn.

%The reason for a player incapable of planning in advance is because he does not have control on the other player's actions. This shows a difference between a plan and a strategy. In fact, the strategies we consider in this paper give a player  choice of single actions rather than a sequence of actions. In this sense, there is no planning in advance. However, it is allowed to think forward. The following strategy is valid:

\item {\it Thinking forward\/:}

$
S^2((\bigcirc (\neg does(a^1_1)) \rightarrow \bigcirc \bigcirc does(a^2_1))\\ 
\begin{array}{ll}
= &\{(0,1,\mybox, \mybox, \mybox, \myboxtimes), a^2_2), (0,1,\mybox, \mybox, \mybox, \myboxtimes), a^2_3),\\
&\ (0,1,\mybox, \mybox, \myboxtimes, \mybox ), a^2_2),(0,1,\mybox, \mybox, \myboxtimes, \mybox ), a^2_4), \\
&\  (0,1,\mybox,\myboxtimes, \mybox, \mybox), a^2_3), (0,1,\mybox,\myboxtimes, \mybox, \mybox), a^2_4)\}
\end{array}
$

This is a strategy for player 2 to try to find an action so that ``{\it he can fill box~1 in his next turn as long as player 1 will not fill it beforehand.}'' Note that $S^2(\bigcirc \bigcirc does(a^2_1)) = \emptyset$ because he can do nothing to guarantee that $does(a^2_1)$ is doable in his next turn.

\item {\it Try anything to win the game within one step:}

$S^1(\bigcirc wins(1))\\
\begin{array}{ll}
= &\{(1,0,\myboxtimes, \mybox, \myboxdot, \mybox), a^1_2), (1,0,\myboxtimes, \mybox, \mybox, \myboxdot), a^1_2), \\
&\ (1,0,\mybox,\myboxtimes, \myboxdot, \mybox), a^1_1), (1,0,\mybox, \myboxtimes, \mybox, \myboxdot), a^1_1), \\
&\ (1,0,\mybox,\myboxtimes, \mybox, \myboxdot), a^1_3), (1,0,\myboxdot, \mybox,\myboxtimes, \mybox), a^1_2), \\
&\ (1,0,\mybox, \mybox, \myboxtimes, \myboxdot), a^1_2), (1,0,\myboxdot, \mybox, \mybox, \myboxtimes), a^1_3), \\
&\ (1,0, \mybox,\myboxdot, \mybox,\myboxtimes), a^1_3)\}
\end{array}
$

This is a simple strategy for player 1 that aims to try any available action to win the game in one step.    
\end{enumerate}
We have seen from the above examples that even though a strategy rule is written in the syntax of (temporal) propositional formulas, its semantics is significantly different from propositional logic therefore we must be very cautious when we start to use the language to design game strategies.

To complete this section, we present a property of our strategy representation for later use.
\begin{Lemma}\label{le:propertiesofSofphi} 
Given a state transition model $M$, for each player~$i$,
\begin{enumerate}
\item $S^i(\varphi_1\wedge\varphi_2)=S^i(\varphi_1)\cap S^i(\varphi_2)$

\item $S^i(\varphi_1)\cup S^i(\varphi_2)\subseteq S^i(\varphi_1\vee\varphi_2)$
\end{enumerate}
\end{Lemma}
\Proof{To prove (1), assume that $(w,a)\in S^i(\varphi_1\wedge\varphi_2)$. It follows that $a\in A^i$ and $M\models_{(w,a)} \varphi_1\wedge \varphi_2$. Then for each reachable path $\delta\in (w,a)^\leadsto$, we have $M,\delta\models \varphi_1\wedge \varphi_2$, which implies $M,\delta\models \varphi_1$ and $M,\delta\models \varphi_2$. It turns out that $M\models_{(w,a)} \varphi_1$ and  $M\models_{(w,a)} \varphi_2$. We get $(w,a)\in S^i(\varphi_1)$ and $(w,a)\in S^i(\varphi_2)$. The other direction is similar.

To prove (2), assume that $(w,a)\in S^i(\varphi_1)\cup S^i(\varphi_2)$. Without loss of generality, suppose that $(w,a)\in S^i(\varphi_1)$, which implies $a\in A^i$ and $M\models_{(w,a)} \varphi_1$. Hence, $M,\delta\models \varphi_1$ for each reachable path $\delta\in (w,a)^\leadsto$. It follows that $M,\delta\models \varphi_1\vee \varphi_2$. Thus we yield $M\models_{(w,a)} \varphi_1\vee \varphi_2$. Given that $a\in A^i$, we conclude $(w,a)\in S^i(\varphi_1\vee \varphi_2)$.}\\

%%%%%%%%%%%%%%%%%%%%%%%%%%%%%%%%%%%%%%%%%%%%%%%%%%%%%%%%%%%%%%%%
\section{Strategy Composition} \label{s:combining}
As mentioned in the introduction, the main motivation of this work is to introduce a formal logical language for defining, comparing and combining strategies. We have provided our formal definition of strategies in both syntactical and semantical levels in the previous sections.  
To facilitate the composition of strategies, in this section we extend our language by two specific connectives, called {\em prioritised disjunction\/} and {\em prioritised conjunction\/}, respectively.

\subsection{Prioritised Disjunction and Conjunction}

The idea behind these two new connectives is the following. The prioritised disjunction ``$\triangledown$'' extends the choice of actions such that if a first strategy fails to apply then a second one offers more options, and if that fails too then a third strategy may offer more options still, and so on. Conversely, the prioritised conjunction ``$\!\vartriangle\!$'' narrows down the choice of actions: if a first strategy allows too many options, then a second strategy may be used to constrain these options, a third strategy may narrow down the options even further, and so on---up to the point where the next strategy in line would lead to empty option.

%The idea was inspired by Brewka {\it et al.}'s Qualitative Choice Logic~\cite{BrewkaBB04}.

\begin{Definition}\label{de:strategyrule}
The set of \emph{strategy rules} is the smallest set such that
\begin{enumerate}
\item a formula in $\mathcal{L}$ is a strategy rule;
 
\item if $r_1,\ldots, r_m$ are strategy rules, then so is $r_1 \triangledown  r_2 \triangledown  \cdots \triangledown r_m$;
%where $m\ge 1$ and $\varphi_i\in \mathcal{L}$ ($1\le i\le m$). 

\item  if $r_1,\ldots,r_m$ are strategy rules, then so is $r_1$$\vartriangle$$r_2$$\vartriangle \cdots$$\vartriangle$$r_m$.
\end{enumerate}
\end{Definition}

Note that the new strategy connectives are introduced as macros rather than as additional connectives in the language $\mathcal{L}$. This is so because we do not want to allow the nesting of the strategy connectives with logical connectives (while nested strategy rules are allowed). For instance, $(\varphi_1\vartriangle\psi_1)\triangledown (\varphi_2\vartriangle\psi_2)$ is a syntactically correct strategy rule while $\varphi\rightarrow (\psi_1\triangledown\!\ \psi_2)$ is not.

\subsection{Semantics of Strategy Rules}

Given a strategy $S$, we let $S$$\upharpoonright_w=\{(w,a): (w,a)\in S\}$, i.e., the set of all the moves at state $w$ specified by $S$. %$S$$\upharpoonright_w=\{(w',a)\in S: w'=w\}$. 

\begin{Definition}\label{de:strategyInterpretation}
Let $r$ be a strategy rule. We define $S^i(r)$ recursively on the structure of $r$ as follows:
\begin{enumerate}
\item If $r=\varphi\in \mathcal{L}$, then $S^i(r)=S^i(\varphi)$.

\item If $r=r_1 \triangledown r_2 \triangledown \cdots \triangledown r_m$, then $(w,a)\in S^i(r)$ iff there exists $k$ (1$\,\leq k\leq m$) such that
  $ (w,a)\in S^i(r_k)\ \mbox{and}\ \bigcup_{j<k}S^i(r_j)\upharpoonright_w=\emptyset. $

\item If $r=r_1$$\vartriangle$$r_2$$\vartriangle \cdots \vartriangle$$r_m$, then $(w,a)\in S^i(r)$ iff there exists $k$ (1$\,\leq k\leq m$) such that
\begin{enumerate}
\item  $ (w,a)\in \bigcap_{j\leq k}S^i(r_j)$ and

\item $k=m$ or $\bigcap_{j\leq k+1}S^i(r_j)\upharpoonright_w=\emptyset$.
\end{enumerate}
\end{enumerate}  
\end{Definition}
Intuitively, $r_1\triangledown r_2\triangledown \cdots \triangledown r_m$ represents a strategy that combines strategies $r_1,r_2,\ldots,r_m$ in such a way that a strategy rule $r_k$ becomes applicable only if none of the higher prioritised rules $r_j (j<k)$ is applicable.
 $r_1$$\vartriangle$$r_2$$\vartriangle$$\cdots $$\vartriangle$$r_m$ tries to apply as many strategy rules all together as possible but gives higher priority to the left rules than the right rules if conflicts occur.
  
The following observations show that the connective $\triangledown$ is indeed a kind of prioritised disjunction and the connective $\vartriangle$ is indeed a kind of prioritised conjunction.
\begin{Lemma}\mbox{}
Given a state transition model $M$, for each player $i$, 
\begin{enumerate}
\item $S^i(r_1)\subseteq S^i(r_1\triangledown r_2)$

 \item $S^i(r_1\vartriangle r_2)\subseteq S^i(r_1)$

 \item If $S^i(r_1)=\emptyset$, then $S^i(r_1\ \triangledown\ r_2) = S^i(r_2)$.
 
\item If $S^i(r_2)=\emptyset$, then $S^i(r_1\vartriangle r_2) = S^i(r_1)$.

\item $S^i(\varphi_1\triangledown (\varphi_1\wedge \varphi_2))=S^i(\varphi_1)$

\item $S^i((\varphi_1\wedge \varphi_2) \triangledown\varphi_2) \subseteq S^i(\varphi_2)$

\item $S^i(\varphi_1\vartriangle (\varphi_1\vee \varphi_2))=S^i(\varphi_1)$

%\item $S^i((\varphi_1\vee \varphi_2)\vartriangle \varphi_2)=S^i(\varphi_2)$
% the other direction is incorrect because S(\varphi) can be empty.

\item $S^i(\varphi_2) \subseteq S^i((\varphi_1\vee \varphi_2)\vartriangle \varphi_2)$

\end{enumerate}
\end{Lemma}
\Proof{
The proof of (1)-(4) is straightforward from Definition~\ref{de:strategyInterpretation}.

To show (5), let $(w,a)\in S^i(\varphi_1\triangledown (\varphi_1\wedge \varphi_2))$. If $(w,a)\not\in S^i(\varphi_1)$, then by Definition~\ref{de:strategyInterpretation}, $(w,a)\in S^i(\varphi_1\wedge \varphi_2)$. By Lemma~\ref{le:propertiesofSofphi} (1), this implies $(w,a)\in S^i(\varphi_1)$---a contradiction. Hence, $(w,a)\in S^i(\varphi_1)$. We have proved $S^i(\varphi_1\triangledown (\varphi_1\wedge \varphi_2))\subseteq S^i(\varphi_1)$. The other direction follows (1).

To show (6), let $(w,a)\in S^i((\varphi_1\wedge \varphi_2) \triangledown \varphi_2)$. Suppose that  $(w,a)\not\in S^i(\varphi_1\wedge \varphi_2)$, then by Definition~\ref{de:strategyInterpretation}, we have $(w,a)\in S^i(\varphi_2)$, as desired. If, on the other hand, $(w,a)\in S^i(\varphi_1\wedge \varphi_2)$, then by Lemma~\ref{le:propertiesofSofphi} (1) we also have $(w,a)\in S^i(\varphi_2)$. 

To show (7), assume that $(w,a)\in S^i(\varphi_1)$. By Lemma~\ref{le:propertiesofSofphi} (2), $(w,a)\in S^i(\varphi_1\vee \varphi_2)$. Hence, $(w,a)\in S^i(\varphi_1)$$\upharpoonright_w \cap\, S^i(\varphi_1\vee \varphi_2)$$\upharpoonright_w $, which implies $(w,a)\in S^i(\varphi_1\vartriangle (\varphi_1\vee \varphi_2))$. The other direction follows (2).

To show (8),
%assume that $(w,a)\in S^i((\varphi_1\vee \varphi_2) \vartriangle \varphi_2)$. Then  $S^i(\varphi_1\vee \varphi_2)$$\upharpoonright_w\not =\emptyset$. By Definition~\ref{de:strategyInterpretation}, we have  $(w,a)\in S^i(\varphi_2)$. Conversely,
assume that $(w,a)\in S^i(\varphi_2)$. By Lemma~\ref{le:propertiesofSofphi} (2), $(w,a)\in S^i(\varphi_1\vee \varphi_2)$. It follows that  $(w,a)\in S^i(\varphi_1\vee \varphi_2)$$\upharpoonright_w \cap\, S^i(\varphi_2)$$\upharpoonright_w$.  By Definition~\ref{de:strategyInterpretation}, we have $(w,a)\in S^i((\varphi_1\vee \varphi_2)\vartriangle\varphi_2)$. 
\comment{The other direction is similar.}}\\

Note that in the above lemma, only one direction of inclusion in items (6) and (8) holds. The other direction does not. For instance, assume that $S^i(\varphi_1)=\{(w,a)\}$ and $S^i(\varphi_2)=\{(w,a),(w,b)\}$. Then $S^i(\varphi_1\wedge \varphi_2)=\{(w,a)\}$. It turns out that $S^i((\varphi_1\wedge\varphi_2)\triangledown \varphi_2)=\{(w,a)\}$. This shows that $S^i(\varphi_2) \not\subseteq S^i(\varphi_1\wedge\varphi_2)\triangledown \varphi_2)$. Furthermore, if $S^i(\varphi_1)=\{(w,a)\}$ and $S^i(\varphi_2)=\emptyset$. By Lemma~\ref{le:propertiesofSofphi}, we have $(w,a)\in S^i(\varphi_1\vee \varphi_2)$. Thus $(w,a)\in S^i((\varphi_1\vee \varphi_2)\vartriangle\varphi_2))\not=\emptyset$, which means that the other direction of item (8) does not hold.

The following lemma shows that the prioritised disjunction can be reduced to binary connectives. Interestingly, the prioritised conjunction does not have such a nice property.

\begin{Lemma} Given a state transition model $M$, for each player $i$, 
$$S^i(r_1\triangledown r_2\triangledown\cdots\triangledown r_m)=S^i(r_1\triangledown (r_2\triangledown\cdots\triangledown r_m)) =S^i((r_1\triangledown\cdots\triangledown r_{m-1})\triangledown r_m)$$
\end{Lemma}
\Proof{We will prove the first equation, the other one will be quite similar. We consider two cases. If $S^i(r_1)$$\upharpoonright_w=\emptyset$ then
	\[ \begin{array}{ll}
	  & (w,a)\in S^i(r_1\triangledown r_2\triangledown\cdots\triangledown r_m)\\
	  \mbox{iff} & (w,a)\in S^i(r_2\triangledown\cdots\triangledown r_m) \\
	  \mbox{iff} & (w,a)\in S^i(r_1\triangledown(r_2\triangledown\cdots\triangledown r_m))
	\end{array} \]
	
On the other hand, if $S^i(r_1)$$\upharpoonright_w\neq\emptyset$ then
	\[ \begin{array}{ll}
	  & (w,a)\in S^i(r_1\triangledown\cdots\triangledown r_m) \\
	  \mbox{iff} & (w,a)\in S^i(r_1) \\
	  \mbox{iff} & (w,a)\in S^i(r_1\triangledown(r_2\triangledown\cdots\triangledown r_m))
	\end{array} \]
}

We remark that the prioritised conjunction cannot likewise be reduced to binary connectives. In general
neither $S^i(r_1$$\vartriangle$$r_2\vartriangle\cdots\vartriangle$$r_m) = S^i((r_1$$\vartriangle \cdots \vartriangle$$r_{m-1})$$\vartriangle $$r_m)$ nor $S^i(r_1$$\vartriangle$$r_2\vartriangle\cdots\vartriangle$$r_m) = S^i(r_1$$\vartriangle$($r_{2}$$\vartriangle \cdots $$\vartriangle $$r_m))$ is true. 
For example, let $S^i(r_1)=\{(w,a_1),(w,a_2)\}$; $S^i(r_2)=\emptyset$; and $S^i(r_3)=\{(w,a_1),(w,a_3)\}$. Then $S^i(r_1$$\vartriangle$$r_2 $$\vartriangle$$r_3)=S^i(r_1)=\{(w,a_1),(w,a_2)\}$. But $S^i((r_1$$\vartriangle$$r_2$)$\vartriangle$$r_3)=S^i(r_1$$\vartriangle$$r_3)=\{(w,a_1)\}$. On the other hand, if $S^i(r_2)=\{(w,a_2),(w,a_3)\}$ and $S^i(r_1)$ and $S^i(r_3)$ are as before, then we get $S^i(r_1$$\vartriangle$$r_2$$\vartriangle$$r_3)=\{(w,a_2)\}$ while $S^i(r_1$$\vartriangle$$(r_2$$\vartriangle$$r_3))=\{(w,a_1),(w,a_2)\}$.

As a remedy, we may define a prioritised conjunction by the prioritised disjunction as follows:
$$\varphi\dot{\vartriangle}\psi =_{def} (\varphi\wedge\psi)\triangledown\varphi$$
Then we extend it to the multi-argument version in the following way: 
$$\varphi_1\dot{\vartriangle}\varphi_2 \dot{\vartriangle}\cdots \dot{\vartriangle}\varphi_m =_{def} ((\varphi_1 \dot{\vartriangle}\varphi_2)\dot{\vartriangle}\cdots) \dot{\vartriangle}\varphi_m$$
Unfortunately the semantics of $\dot{\vartriangle}$ coincides with $\vartriangle$ in the binary version but not in the multi-argument version.  We will investigate this alternative in our future work.

\subsection{Complete Strategies and Deterministic Strategies}
When we build a software game player, we need to instruct it what to do in each possible situation. In other words, each player should be equipped with a functional strategy. In the following, we demonstrate how to generate a complete and/or deterministic strategy by using our prioritised connectives.
 
We say that a strategy rule is \emph{consistent} for player $i$ if it represents a valid strategy; a rule is \emph{complete} for~$i$ if it represents a complete strategy for~$i$; and \emph{deterministic} for~$i$ if it represents a deterministic strategy for~$i$. Similarly, a strategy rule is {\it functional} if it is complete and deterministic. Note that all these concepts are player-specific.

The following theorems show a number of nice properties of the prioritised connectives, which give us a guideline for how to design a strategy with desired properties. The first result deals with consistency of strategy rules.

\begin{Theorem}
Given a state transition model $M$,
\begin{enumerate}
\item  $r_1 \triangledown \cdots \triangledown r_m$ is consistent if and only if there is a $k$ (1$\,\leq k\leq m$) such that $r_k$ is consistent.

\item $r_1$$\vartriangle\cdots\vartriangle$$r_m$ is consistent if and only if $r_1$ is consistent.
\end{enumerate}
\end{Theorem}
\Proof{ To prove (1), let $r = r_1 \triangledown \cdots \triangledown r_m$.
Assume that $S^i(r)$ is non-empty for player $i$, then there is $(w,a)\in S^i(r)$ such that $(w,a)\in S^i(r_{k})$ for some $k$, which means that $r_k$ is consistent for player $i$. Conversely, let $k$ be the smallest number such that $S^i(r_k)$ is non-empty. This implies that $S^i(r_1 \triangledown \cdots \triangledown r_m) = S^i(r_k \triangledown \cdots \triangledown r_m)$. By the above two lemmas we have $S^i(r_k)\subseteq S^i(r_1 \triangledown \cdots \triangledown r_m)$, which means that $r$ is consistent.

To prove (2), we also let $r= r_1$$\vartriangle\cdots\vartriangle$$r_m$. Obviously if $r_1$ is inconsistent, so is $r$.  Assume that $r_1$ is consistent. Then there is $(w,a)\in S^i(r_1)$. Let $k$ be the biggest number such that $\bigcap_{j\leq k}(S^i(r_j)\upharpoonright_w)\not =\emptyset$. Obviously $k\geq 1$. By Definition~\ref{de:strategyInterpretation} we have $\bigcap_{j\leq k}(S^i(r_j)\upharpoonright_w) \subseteq S^i(r)\upharpoonright_w$. Thus $r$ is consistent. 
}\\

The second result shows us how to generate a complete or deterministic strategy.

\begin{Theorem} \label{pr:pdisjunction}
Given a state transition model $M$, for each player $i$
\begin{enumerate}
\item If $r_1$ or $r_2$ is complete, so is $r_1 \triangledown r_2$.

\item If $r_1$ is complete, so is $r_1$$\vartriangle\cdots\vartriangle$$r_m$.

\item If $r_1$ and $r_2$ are deterministic, so is $r_1 \triangledown r_2$.

\item If $r_1$ is deterministic, so is $r_1$$\vartriangle\cdots\vartriangle$$r_m$.
\end{enumerate}
\end{Theorem}
\Proof{
(1) and (4) are straightforward from Definition~(\ref{de:strategyInterpretation}).

To show (2), assume that $r_1$$\vartriangle\cdots\vartriangle$$r_m$ is incomplete. Then there exists a reachable state $w\in W\cap\mathcal{P}(G)$ such that $S^i(r_1$$\vartriangle\cdots\vartriangle$$r_m)$$\upharpoonright_w = \emptyset$. By Definition~\ref{de:strategyInterpretation}, this can happen only if  $S^i(r_1)$$\upharpoonright_w = \emptyset$, which contradicts the assumption that $r_1$ is complete.

To prove (3), assume that $(w,a),(w,a')\in S^i(r_1\triangledown r_2)$. If $S^i(r_1)$$\upharpoonright_w\not=\emptyset$, we have $(w,a),(w,a')\in S^i(r_1)$ according to Definition~\ref{de:strategyInterpretation}. Since $r_1$ is deterministic, this implies $a=a'$.  On the other hand, if $S^i(r_1)$$\upharpoonright_w = \emptyset$,  then we have  $(w,a),(w,a')\in S^i(r_2)$$\upharpoonright_w$, which also implies $a=a'$ since $r_2$ is deterministic.
}\\

Statement (1) in the above theorem provides us with an easy way of generating a complete strategy: {\it create a trivial complete strategy first and then combine it with other strategies using the prioritised disjunction}. Note that creating a trivial complete strategy is rather easy: {\it let the agent do anything available}.\footnote{In this case, we might have to give the trivial strategy the lowest priority in the disjunction.} Statement (2) tells us that once we get a complete strategy, we can further refine the strategy targeting more specific properties, say deterministic thus functional, using the prioritised conjunction without losing its completeness. Statement (3) shows us another feasible way of generating a functional strategy: instead of creating a complete strategy then refine it into a deterministic one, we can {\it devise a set of specific deterministic strategies first and then combine them with the prioritised disjunction targeting a complete strategy}. Example \ref{ex:boxdot5} demonstrates how the above mentioned approaches can be applied to the CrossDot game. Before doing that, let's show another nice property of the prioritised connectives.

\begin{Theorem}\label{pr:connectiveproperties1}
Given a state transition model $M$, for each player $i$ 
\begin{enumerate}
\item If $r_1$ is complete, then $S^i(r_1 \triangledown \cdots \triangledown r_m) = S^i(r_1)$.

\item If $r_1$ is deterministic, then $S^i(r_1$$\vartriangle\cdots\vartriangle$$r_m) = S^i(r_1)$.
\end{enumerate}
\end{Theorem}

\Proof{ To prove (1), we only have to show $ S^i(r_1 \triangledown \cdots \triangledown r_m)\subseteq S^i(r_1)$. Assume that $(w,a)\in S^i(r_1 \triangledown \cdots \triangledown r_m)$. Because $r_1$ is complete, $S^i(r_1)$$\upharpoonright_w\not=\emptyset$. By Definition~\ref{de:strategyInterpretation} we have  $(w,a)\in S^i(r_1)$.

To show (2), we need to show that $S^i(r_1)\subseteq S^i(r_1$$\vartriangle\cdots\vartriangle$$r_m)$. If
$(w,a)\in S^i(r_1)$, then $S^i(r_1)$$\upharpoonright_w=\{(w,a)\}$ since $r_1$ is deterministic. By Definition~\ref{de:strategyInterpretation} it follows that $(w,a)\in S^i(r_1$$\vartriangle\cdots\vartriangle$$r_m)$.}\\

Combining the two statements of this theorem, we can say that if $r_1$ is functional, then $S^i(r_1 \triangledown \cdots \triangledown r_m)$$=$$S^i(r_1$$\vartriangle\cdots\vartriangle$$r_m)$ $=$$S^i(r_1)$. This means that once a strategy is functional, neither extending nor refining it with strategies of lower priority has any effect. In fact, our goal of introducing the prioritised connectives is to facilitate the design of functional strategies. Once a functional strategy has been obtained, these connectives automatically stop working.

%\begin{ContinueExample}
\begin{Example}\label{ex:boxdot5}\rm
% the game in Example~\ref{ex:boxdot1}. 
Consider the CrossDot game scenario in Example 1. We define a few strategy rules for player~$i$ as follows: 
\begin{itemize}
\item Fill a box next to a box that contains player $i$'s mark:
\begin{equation} 
\begin{array}{ll}
   fill\_next^i = & \bigvee_{1< j\leq m}(\neg p^1_j \wedge\neg p^2_j\wedge  p^i_{j-1} \wedge does(a^i_{j})) \ \vee\\ 
  &   \bigvee_{1\leq j<m} (\neg p^1_j\wedge \neg p^2_j \wedge p^i_{j+1} \wedge does(a^i_{j}))
 \end{array}
\end{equation}

\item Fill an isolated box (i.e., whose immediate neighbours  are empty):
\begin{equation}
\begin{array}{ll}
\fillisolated^i\,  = \bigvee_{1< j< m}(\!\!\! & (\neg p^1_{j-1}\wedge \neg p^2_{j-1})\wedge (\neg p^1_{j+1}\wedge \neg p^2_{j+1})\!\!\! \\
& \wedge\,(\neg p^1_{j}\wedge \neg p^2_{j}) \wedge does(a^i_{j}))
  \end{array}
\end{equation}

\item Fill any empty box:
\begin{equation}
 \begin{array}{l}
\fillany^i\,=\,\bigvee_{1\leq j\leq m}(\neg p^1_j\wedge \neg p^2_j\wedge does(a^i_j))
\end{array} 
\end{equation}

\item Try $fill\_next^i$ first. If this fails, try $\fillisolated^i$, then try $\fillany^i$:
\begin{equation}\label{eq:combined}
\combined^i \,=\, fill\_next^i\ \triangledown\ \fillisolated^i\ \triangledown\ \fillany^i
\end{equation}
\end{itemize}
\end{Example}
%\end{ContinueExample}

\begin{Observation}
Let $M=(G,v)$ be the state transition model defined in Observation \ref{ob:first}. The strategy rule $\combined^i$ is complete for player $i$.
\end{Observation}
\Proof{We first prove that the strategy $\fillany^i$ is complete. Assume an arbitrary reachable state $w\in W\cap \mathcal{P}(G)$ such that $l^i(w)\not=\emptyset$ (recall formula~(\ref{eq:lw}) for the definition of~$l^i$ and the definition of completeness of a strategy in Section~\ref{subse:propertiesOfStrategies}). Let $a^i_j\in  l^i(w)$. It follows that $M\models_{(w,a^i_j)}legal(a^i_j)$. By~(\ref{eq:8}), we know that $M\models_{(w,a^i_j)}\neg p^1_j\wedge \neg p^2_j$, hence $M\models_{(w,a^i_j)}\neg p^1_j\wedge \neg p^2_j\wedge does(a^i_j)$, which implies $(w,a^i_j)\in S^i(\fillany^i)$. Therefore $\fillany^i$ is a complete strategy rule for player $i$. According to Theorem~\ref{pr:pdisjunction} (1), we know that the strategy rule $\combined^i$ is also complete. 
}\\

A complete strategy for a player gives the player a feasible option (if any) for any given situation. However, a software game player could still be unsure about what to do under a complete strategy if there is more than one option in the same situation.
%In this case we have to tell the software player what  option to take.
The prioritised conjunction provides us with a way to narrow down multiple options.

%\begin{ContinueExample}\rm
\begin{Example}\rm\label{Ex:ccc}
For our running example game, let $c^i_t=\bigvee\limits_{j= 1}^{t} does(a^i_j)$ where $1\leq t\leq m$ (with $m$ being the overall number of boxes as usual), then  $c^i_t$ represents the strategy of $i$ to place an object in any box between $1$ and a given number $t$. Let
\comment{
\begin{equation}\label{eq:fill_leftmost}
\fillleftmost^i = c^i_m \vartriangle\cdots \vartriangle c^i_1
\end{equation}
which formalises the strategy of filling the leftmost empty box. Let}
\begin{equation}\label{eq:thoughtfull}
\thoughtful^i = \combined^i \vartriangle c^i_m \vartriangle\cdots \vartriangle c^i_1
\end{equation}
where $\combined^i$ is the strategy rule defined in Example~\ref{ex:boxdot5} (cf.~(\ref{eq:combined})). 
\end{Example}

It is not hard to see that if $\combined^i$ gives more than one boxes to fill, $thoughtful^i$ will choose the left most one.  

\begin{Observation}
Let $M=(G,v)$ be the state transition model defined in Observation \ref{ob:first}. The strategy rule $\thoughtful^i$ is functional for player $i$.
\end{Observation}
\Proof{By Theorem~\ref{pr:pdisjunction} (2), $\thoughtful^i$ is complete because $\combined^i$  is complete.
 We prove that $\thoughtful^i$ is deterministic. Assume that $$(w,a^i_{j_1}), (w,a^i_{j_2})\in S^i(\thoughtful^i)$$ such that, without loss of generality, $j_1\leq j_2$.  Let $t$ be the smallest number (between $1$ and $m$) such that $H=S^i(\combined^i)$$\upharpoonright_w$$\cap \bigcap\limits_{j=t}^mS^i(c^i_j)$$\upharpoonright_w$ is not empty. Such a $t$ exists because $S^i(\combined^i)$$\upharpoonright_w$$\cap S^i(c^i_m)$$\upharpoonright_w\not = \emptyset$. By Definition~\ref{de:strategyInterpretation}, we have $(w,a^i_{j_1})\in H$ and $(w,a^i_{j_2})\in H$, hence  $(w,a^i_{j_1}), (w,a^i_{j_2})\in S^i(c_t^i)$$\upharpoonright_w$. Hence $ j_1\leq j_2\leq t$. If $t=1$, then $a^i_{j_1}=a^i_{j_2}=a^i_1$.
Otherwise, i.e.\ if $t>1$, then $S^i(\combined^i)$$\upharpoonright_w\cap \bigcap\limits_{j=t-1}^{m}S^i(c^i_j)$$\upharpoonright_w = \emptyset$, which implies that neither $(w,a^i_{j_1})$ nor $(w,a^i_{j_2})$ belongs to $S^i(c_{t-1}^i)$$\upharpoonright_w$.  Note that $S^i(c_{t-1}^i)$$\upharpoonright_w = \{(w,a^i_j)\in \mathcal{P}(G):j\leq t-1\}$. Thus $t\leq j_1\leq j_2$. By the above assumption, we have $j_1= j_2=t$.
We have proved that  $\thoughtful^i$ is deterministic, thus it is functional.}\\
%\end{ContinueExample}

To summarise, the prioritised connectives provide a natural way of refining a strategy. If a strategy is too restricted in that it can only be applied to few states, it can be extended using prioritised disjunction. If, on the other hand, a strategy is too generic in that it leaves too many options, it can be strengthened using prioritised conjunction. Once a strategy has been functional, further extension or refinement using the prioritised disjunction or conjunction take no effect as long as give the existing strategy the highest priority. 

%%%%%%%%%%%%%%%%%%%%%%%%%%%%%%%%%%%%%%%%%%%%%%%%%%%%%%%%%%%%%%%%
\section{Reasoning About Strategies} \label{se:reasoning}

Knowing how to write strategies for a game-playing agent, we now consider the question whether a strategy meets its goal, for instance, if it is guaranteed to lead to a winning state or to a desirable state given the strategies that the other players use. In this section, we will demonstrate by using our running example how to reason about strategies within our framework. We also demonstrate how to design a strategy using the prioritised connectives to meet desired properties.

%Instead of developing a proof theory for this purpose, we will show how to verify if a strategy can bring about an expected outcome with a model-based approach.

\subsection{Compliance with a strategy}
In order to verify whether a strategy can bring about an expected result for a player in a game, we assume that the player complies with the strategy all the way through the game and observes the outcome of the game. 

Let $M=(G,v)$ be a state transition model of a game $G$ and $S$ a strategy of the game for player $i$. We say that $M$ (or $G$) {\it complies with $S$ by player} $i$ if for each reachable move $(w,a)$ of player $i$ in $G$, $(w,a)\in S$. In other words, player $i$ follows the strategy $S$ whenever he makes a move.\footnote{We may understand the concept in the following way. Given an arbitrary state transition model $M=(G,v)$ and a strategy $S$ of player $i$ in $M$, let $M'$ be another state transition model that is exactly the same as $M$ except for the legality of actions for player $i$ in such a way that the legality relation $l'$ in $M'$ is $l\cap S$, where $l$ is the legality relation in $M$. We then view $M'$ as the reduction of $M$ once player $i$ complies with strategy $S$.}

The following observation shows that for any CrossDot game when $k=2$ and $m>2$, Player 1 wins as long as he plays the strategy $\thoughtful^i$ all the way through the game.

\begin{Observation}
Let $M$ be a state transition model for the CrossDot game with $k=2$ and $m>2$. Assume that player $i$ takes the first turn. If $M$ complies with the strategy $S^i(\thoughtful^i)$ by player $i$, then
$$M\models terminal\rightarrow wins(1)$$
In other words, player $i$ wins as long as he takes the first turn and follows the strategy rule $\thoughtful^i$.
\end{Observation}
\Proof{Without loss of generality, we assume that $i=1$.  We have to show that $M,\delta\models terminal\rightarrow wins(i)$ for any reachable path $\delta$. Since the definitions of $terminal$ and $wins$ do not contain $does$, $legal$ and $\bigcirc$ (cf.\ formulas~(\ref{eq:3}) \& (\ref{eq:4})), we only need to show for any reachable terminal state $w_e$ (which is a special case of a reachable path)  $M,w_e\models wins(1)$.  
To this end, we assume a complete path $\delta = \pathe$, where $w_0=\bar{w}$ and $w_e\in t$. Since player $1$ has the first turn and plays with strategy $\thoughtful^1$, the action the player takes in the initial state must be $a^1_2$, i.e., $a_0=a^1_2$. If player 2 responds with action $a^2_1$, i.e., $a_1=a^2_1$, player 1 will then take action $a^1_3$ and win the game at state $w_3$. If player 2 responds with any other action, player 1 will take $a^1_1$ and also will win the game. In any case, $\delta$ ends up with $w_e=w_3$ in which player 1 wins. Therefore $M,w_e\models wins(1)$. We  conclude $M\models terminal\rightarrow wins(1)$.
}\\

The strategy $\thoughtful$ seems like a ``smart'' strategy that can guarantee a winning state for the player who takes the first turn when $k=2$. However it is less mighty when taken by the second player, in which case the strategy cannot even compete the following ``trivial'' strategy 
\begin{equation}\label{eq:fillleft}
\fillleftmost^i=c^i_m \vartriangle\cdots \vartriangle c^i_1
\end{equation}
where $c^i_j$ was defined in Example~\ref{Ex:ccc}.   

\begin{Observation}\label{ob:secondlost}
Let $M$ be a state transition model for a CrossDot game with $k=2$ and $m>2$. If $M$ complies with  $S^1(\fillleftmost^1)$ by Player 1 and with $S^2(\thoughtful^2)$ by Player 2, then 
$$M\models terminal \rightarrow wins(1)$$
\end{Observation}
\Proof{ It is not hard to prove that $\fillleftmost^1$ is functional. Since $\thoughtful^2$ is also functional, there is only one complete path in the game: 
$$(1,0, \mybox, \mybox, \mybox, \mybox, \cdots, \mybox)\stackrel{a^1_1}{\rightarrow}(0,1, \myboxtimes, \mybox, \mybox, \mybox,\cdots, \mybox)$$$$\stackrel{a^2_3}{\rightarrow}(1,0, \myboxtimes, \mybox, \myboxdot, \mybox,\cdots, \mybox)\stackrel{a^1_2}{\rightarrow}(0,1, \myboxtimes, \myboxtimes, \myboxdot, \mybox,\cdots, \mybox)$$ 
which implies that player 1 takes the left most box, followed by player 2 fills an isolated box and finally player 1 fills the second box and wins. We can easily verify that $terminal \rightarrow wins(1)$ is valid with any segment of the path in $M$.}\\

The failure of $\thoughtful$ is not because it is not ``smart'' enough but because the $\thoughtful^2$ strategy requires to fill an isolated box at the first move, which provides the first player with a chance to win. 

%A player should protect himself before making an passive strategy. 

\subsection{Reasoning About Other Players' Strategies}

The examples of strategies we have shown up to now are all from a single player's viewpoint, which is obviously not sufficient. We should also reason about other players' strategies.

From Observation~\ref{ob:secondlost} we learnt that a player should check for existing threats before applying any ``aggressive'' strategy, like $\thoughtful$. The following formula defines a {\it defence} strategy, which says that if my opponent can win by filling box~$j$ at next step, then I should mark it now to prevent an immediate loss: 
\comment{
\begin{equation}\label{eq:defence}
\begin{array}{l}
\defence^i\, = \\
\ \ (\textstyle\bigwedge^m_{j=1}(\bigcirc (does(a^{-i}_j)\wedge \bigcirc wins(-i))\rightarrow does(a^i_j)))\!\vartriangle\! c^i_m \vartriangle\cdots \vartriangle c^i_1
\end{array}
\end{equation}
}
\comment{
\begin{equation}\label{eq:defence}
\begin{array}{ll}
\defence^i\, = & \bigvee\limits^m_{j=1}(does(a^i_j)\wedge \bigwedge\limits^m_{k=1}(\neg \bigcirc (does(a^{-i}_k)\wedge\bigcirc wins(-i))))
\end{array}
\end{equation}
}
%\comment{
\begin{equation}\label{eq:defence}
\begin{array}{l}
\defence^i = 
%& (\bigvee^m_{j=1}(\neg does(a^i_j)\wedge \bigcirc does(a^{-i}_j)\wedge \bigcirc\bigcirc wins(-i)))  \\
%& \wedge 
\bigwedge\limits^m_{j=1}(\bigcirc (does(a^{-i}_j)\wedge \bigcirc wins(-i))\rightarrow does(a^i_j))%\!\vartriangle \!\thoughtful^i
\end{array}
\end{equation}
%}
where $-i$ stands for the opponent of $i$. Note that $\defence^i$ is neither deterministic nor complete. To create a functional strategy with defence, we let
\begin{equation}\label{eq:cautious}
\cautious^i=(\defence^i\!\vartriangle\! c^i_m \vartriangle\cdots \vartriangle c^i_1)\ \triangledown\ \thoughtful^i
\end{equation}
Obviously, $\defence^i\!\vartriangle\! c^i_m \vartriangle\cdots \vartriangle c^i_1$ is deterministic. Since $\thoughtful^i$ is functional, therefore $\cautious^i$ is functional.

The following observation shows that if the second player plays $\cautious^2$, which means to protect himself before attacking his opponent, then player~1 cannot win with the $\fillleftmost$ strategy (cf. equation (\ref{eq:fillleft})).

\begin{Observation}\label{ob:secondwin}
Let $M$ be a state transition model for the CrossDot game with $k=2$ and $m>2$. If $M$ complies with $S^1(\fillleftmost^1)$ by Player 1 and with $S^2(\cautious^2)$ by Player 2, then 
$$M\models terminal \rightarrow \neg wins(1)$$
\end{Observation}
\Proof{ 
Since both players' strategies are functional, there is only one complete path in the game. Assume that the complete path is $\delta=\pathe$. The first action taken by player 1 must be  $a_0=a^1_1$. The first two states in the path is then
$$w_0=(1,0, \mybox, \mybox, \mybox, \mybox, \cdots, \mybox) \mbox{ and } w_1=(0,1, \myboxtimes, \mybox, \mybox, \mybox,\cdots, \mybox)$$

To know which action player 2 chooses at state $w_1$, we calculate $S^2(\defence^2)$$\upharpoonright_{w_1}$. 
In fact, we will show that $S^2(defence^2)$$\upharpoonright_{w_1}=\{(w_1,a^2_2)\}$. 
Assume any reachable path $\delta'\in (w_1,a^2_2)^\leadsto$. 
We verify $$M, \delta'\models\bigwedge^m_{j=1}\ (\bigcirc (does(a^{1}_j)\wedge \bigcirc wins(1))\rightarrow does(a^2_j))$$
This is obviously true because if $j=2$, then $does(a^2_j)$ is satisfied. For any $2<j\leq m$, $\bigcirc (does(a^{1}_j)\wedge \bigcirc wins(1))$ is false. 

Next we show that for any $a_1\not=a^2_2$, $(w_1,a_1)\not\in S^2(\defence^2)$$\upharpoonright_{w_1}$. Let $\delta' = w_1 \stackrel{a_1}{\rightarrow}w'_2\stackrel{a^1_2} {\rightarrow}w'_3$. It is easy to verify that 
$$M, \delta'\models \bigcirc (does(a^{1}_2)\wedge \bigcirc wins(1))\wedge \neg does(a^2_2)$$
It follows that
$$M, \delta'\not\models\bigwedge^m_{j=1}\ (\bigcirc (does(a^{1}_j)\wedge \bigcirc wins(1))\rightarrow does(a^2_j))$$
Note that $\delta'\in (w_1,a_1)^\leadsto$. Therefore we have 
$$M\not\models_{(w_1,a_1)}\bigwedge^m_{j=1}\ (\bigcirc (does(a^{1}_j)\wedge \bigcirc wins(1))\rightarrow does(a^2_j))$$
which implies $(w_1,a_1)\not\in S^2(\defence^2)$$\upharpoonright_{w_1}$. This completes the proof that $S^2(\defence^2)\upharpoonright_{w_1}=\{(w_1,a^2_2)\}.$ Hence, for the complete path $\delta$ we find that $a_1=a^2_2$ and $w_2=(1,0, \myboxtimes,\myboxdot,  \mybox, \mybox,\cdots, \mybox)$. The game continues in the same way until all boxes are filled without a winner.
}\\

The instances of the CrossDot game we considered above are limited to the simple case where $k=2$. In the following, we proposed a solution to the game in the general setting where $k$ can be any number larger than $2$ (and $\leq m$). Consider the following strategies:
\begin{itemize}
\item Fill a box next to an opponent's box:
 $$
 \begin{array}{ll}
 \fillonext^i\,= &(\bigvee_{1\leq j<m} (\neg p^1_j\wedge \neg p^2_j \wedge p^{-i}_{j+1} \wedge does(a^i_{j})))\\
& \ \triangledown   (\bigvee_{1< j\leq m}(\neg p^1_j \wedge\neg p^2_j\wedge  p^{-i}_{j-1} \wedge does(a^i_{j})))
 \end{array}
 $$
Note that priority is given to the left empty box if exists.
\item Passive defence:
\begin{equation}\label{eq:aggressivedefence}
% \begin{array}{ll}
\passivedefence^i\!=\! 
((\defence^i\!\vartriangle\!\fillonext^i)\triangledown \fillany^i)\!
 \vartriangle c^i_m\!\vartriangle\!\cdots\!\vartriangle\!c^i_1
% \end{array}
\end{equation}
Note that we use prioritised conjunction, instead of disjunction, to combine $\defence$ and $\fillonext$. This is because $\defence$ gives arbitrary actions whenever there is no immediate loss for player $i$, in which case $\fillonext$ takes over.
\end{itemize}

The following observation shows that such a passive defence strategy guarantees no loss for a player no matter whether it is taken by the first player or the second player, and no matter what the strategy of the other player, for any instance of the CrossDot game with $k>2$ and $m>2$.

\begin{Observation}
Let $M$ be a state transition model for a CrossDot game with $k>2$ and $m\geq k$. If $M$ complies with $S^i(\passivedefence^i)$ by player $i$, then $$M\models terminal\rightarrow \neg win(-i)$$
where $-i$ represents the opponent player of $i$. In other words, a player never loses as long as he plays the $\passivedefence$ strategy.
\end{Observation}
\Proof{Obviously we only have to consider the case when $k=3$ because if a strategy can help a player effectively blocking his opponent to own three consecutive boxes, he can also block his opponent to form any longer line of consecutive boxes. Also if we can prove the strategy to be effective for player 2, it is sufficient to show that it is also effective for player 1 because he can make his first move at random and then copy player 2's no-loss strategy (strategy stealing). Altogether this allows us to restrict the proof to $k=3$ and $i=2$.

It is easy to show that $\passivedefence^2$ is functional.  To show $M\models terminal\rightarrow \neg win(1)$, we only have to verify that for any complete path $\delta=\pathe$, we have $M, w_e\models \neg win(1)$. Recall that $w_e$ is a special reachable path that contains only the terminal state $w_e$. If $w_e$ is a winning state for player 2 or a tie state, we have $M, w_e\models \neg win(1)$. 
 
 Suppose by contradiction that $w_e$ is a winning state of player 1. It then turns out $e$ is an odd number and there are at least three consecutive crosses in some states along the complete path $\delta$. In other words, there must be a cross next to two other crosses. However, we will show by induction on the length of the path that this can never happen. More precisely, we claim that 
{\it in each state $w_{2j} (0\leq j\leq (e-1)/2)$, each box filled with a cross must be either adjacent to a box with a dot or at an end (left or right) of the line, meanwhile the box on its left, if any, must not be empty. 
% (1) ?XO: ? = X or O If ?=X, then its left can never be X, in that case player 1 has won and can never be empty. 
% (2) OX?: ? = X or O or _ If ? = _, then its right can never be X. If ?=X, its right must be a dot
% (3) ]X?:   ? = X or O or _ If ? = _, then its right can never be X. If ?=X, its right must be a dot
% (4) ?X[ :  ? = X or O   If ?=X, then its left can never be X, in that case player 1 has won and can never be empty. 
}
Note that if the claim is true in $w_{e-1}$, then player 1 cannot win in state $w_e$ no matter which action he chooses in state $w_{e-1}$.

Obviously the claim holds when $j=0$ because all boxes are empty in the initial state $w_0$. Now we assume that the claim holds in state $w_{2(j-1)}$ where $j\leq (e-1)/2$. We show that the claim also holds at $w_{2j}$. 

Suppose that the action $a_{2(j-1)}$ that player 1 took in state $w_{2(j-1)}$ was $a^1_v$, i.e., $a_{2(j-1)}=a^1_v$. Obviously this action can at most affect satisfaction of the claim on the box itself or its two immediate neighbours, i.e., $v-1$, $v$ and $v+1$ (if exists). Therefore we only consider the effect of player 1's action $a^1_{v}$ on these three possible boxes. 

If box $v$ is not at the right end, by the induction assumption, box $v+1$ can never be with a cross in state $w_{2(j-1)}$; otherwise box $v$ were not empty thus $a^1_v$ were not doable.  Therefore box $v+1$ (if exists) satisfies the claimed conditions in state $w_{2j}$. 

We then consider satisfaction of boxes $v-1$ and $v$ with three cases:  

i). If box $v-1$ (if exists) has been filled with a dot, both boxes $v-1$ and $v$ satisfy the conditions of the claim, which remain true in state $w_{2j}$. 

ii). If box ${v-1}$ has been already filled with a cross before state $w_{2(j-1)}$, by induction assumption, box $v-1$ is either at the left end or next to a box with a dot on its left. In both situations, box $v-1$ satisfies the claimed conditions that remain true in state  $w_{2j}$. After player 1 fills box $v$ with a cross in state $w_{2(j-1)}$, player 2 must respond with action $a_{2j-1}=a^2_{v+1}$ by applying $\defence^2$ in state $w_{2j-1}$ to prevent an immediate loss unless the box $v$ is at the right end or $v+1$ has already occupied by a dot. In both cases, box $v$ satisfies the claimed conditions in state  $w_{2j}$. Note that the claim guarantees that $\defence^2$ is not applicable to any other boxes. 

iii). If box $v-1$ was empty in state $w_{2(j-1)}$, by the construction of $\passivedefence^i$, player 2 must respond with action $a^2_{v-1}$ either because $v-1$ is an immediate loss position or by applying strategy $\fillonext^2$ after player 1 fills box $v$. After $v-1$ filled with a dot, both boxes $v-1$ and $v$ satisfy the claim in state  $w_{2j}$. Note that $\defence^2\!\vartriangle\!\fillonext^2$ is not applicable to any box other than $v-1$ (note that $\fillonext^2$ gives priority to the left empty box.

We have verified the claim holds in each state $w_{2j}$ $(0\leq j \leq (e-1)/2)$, including state $w_{e-1}$, which implies that the player 2 does not lose in state $w_e$.
}\\

Note that the $\defence$ strategy (cf. equation (\ref{eq:defence})) plays a crucial role in the solution.  Assume that a game gets into the following situation after a few moves:
$$\myboxdot,\myboxtimes,\mybox, \mybox, \myboxdot,\myboxtimes,\myboxtimes, \mybox, \cdots$$
and it is player 2's turn. If player 2 does not have the $\defence$ strategy but simply use $\fillonext^2$ to black the opponent, he would take action $a^2_3$ instead of $a^2_8$, which gives player 1 a chance to win.\\
%: 
%$$\myboxdot,\myboxtimes,\myboxdot, \mybox, \myboxdot,\myboxtimes,\myboxtimes,\myboxtimes, \cdots$$

The strategy $\passivedefence$ can effectively prevent from losing but it is hard to win because it does not encode any winning strategy. The reader is invited to extend the strategy with more aggressive rules so that if the other player is ``not that smart'', it can have a chance to win.   

%%%%%%%%%%%%%%%%%%%%%%%%%%%%%%%%%%%%%%%%%%%%%%%%%%%%%%%%%%%%%%%%
\section{Computing With Strategies} \label{s:computing}

We now turn to the question of how to actually compute with strategy rules. Generally speaking, the conceptual simplicity of our language and the fact that it is not tied to a specific action formalism should make it easy to incorporate knowledge of strategies into various methods for the design and analysis of intelligent agents. To illustrate this, we will adopt here a very general calculus for reasoning about actions, the Situation Calculus (see, e.g., \cite{reiter2001knowledge}), and show how our strategy representation can be easily integrated. Reasoning problems about strategies can take different forms, and we will specifically consider two of them. First, we will show how the calculus can be used to infer the possible outcomes of a game given information about the strategies of all players. Second, we will illustrate how players can reason about the strategies of opponents to infer their best course of action. We will also show how our variant of the Situation Calculus forms the basis for an encoding of game rules and a restricted class of strategies as Answer Set Programs.

\subsection{Example\/: Situation Calculus}

\noindent
The Situation Calculus is a formalism for reasoning about actions and change that is based on classical predicate logic with a few pre-defined language elements\/:
\begin{itemize}
  \item $s_0$, a constant denoting the {\em initial situation\/}; and $\Do(\alpha,\sigma)$, a constructor denoting the situation resulting from doing action~$\alpha$ in situation~$\sigma$;
  \item $\Holds(\varphi,\sigma)$, a predicate denoting that {\em fluent\/}~$\varphi$ (i.e., an atomic state feature) is true in situation~$\sigma$;
  \item $\Poss(\alpha,\sigma)$, a predicate denoting that action~$\alpha$ is possible in situation~$\sigma$.
\end{itemize}
For our purpose, we extend the base language of the Situation Calculus by the two game-specific predicates $\Wins(\iota,\sigma)$ and $\Terminal(\sigma)$ meaning, respectively, that the game is won for player~$\iota$ in situation~$\sigma$ and that $\sigma$ is a terminal game position. With this, any axiom in our game specification language can be easily rewritten for the Situation Calculus similar to an existing mapping of the Game Description Language (GDL) into this calculus~\cite{B:65}. To this end, let $A$ and $S$ be two distinct variables (standing for any action and situation, respectively), then a formula $\varphi$ in our language $\mathcal{L}$ can be translated into a Situation Calculus axiom $\sitc{\varphi}{A}{S}$ by the following inductive definition:
  \[ \begin{array}{rcl}
    \sitc{p}{A}{S} & := & \Holds(p,S) \\
   \sitc{(\neg\varphi)}{A}{S} & := & \neg\sitc{\varphi}{A}{S} \\
   \sitc{(\varphi\wedge\psi)}{A}{S} & := & \sitc{\varphi}{A}{S}\wedge\sitc{\psi}{A}{S} \\
    \sitc{does(a)}{A}{S} & := & A=a \\
    \sitc{legal(a)}{A}{S} & := & \Poss(a,S) \\
    \sitc{wins(i)}{A}{S} & := & \Wins(i,S) \\
    \sitc{(\bigcirc\varphi)}{A}{S} & := & (\forall A')\,\sitc{\varphi}{A'}{\Do(A,S)} \\
    %Michael: What is the range of A'?
    \sitc{init}{A}{S} & := & S=s_0 \\
    \sitc{terminal}{A}{S} & := & \Terminal(S)
  \end{array} \]
  
\begin{ContinueExample}\label{ex:boxdot4}\rm\ 
Recall the specification of the game in Example~\ref{ex:boxdot1}. Applying the construction from above to formulas~(\ref{eq:2})--(\ref{eq:12}) yields the following, after some slight syntactic simplifications.
%\[ \neg[\Holds(p^1_j,S)\wedge \Holds(p^2_j,S)] \]
\[ \neg\Holds(p^i_j,s_0) \]
\[ \Holds(turn(1),s_0)\wedge\neg\Holds(turn(2),s_0) \]
\[ \Wins(i,S)\equiv \bigvee_{j=1}^{m-k+1}\  \bigwedge_{l=j}^{j+k-1} \Holds(p^i_l,S) \]
\[ \begin{array}{ll} \Terminal(S)\equiv\!\! & \Wins(1,S)\vee \Wins(2,S)\,\vee\\ & (\bigwedge_{1\leq j\leq m}(\Holds(p^1_j,S)\vee \Holds(p^2_j,S))) \end{array} \]
\[ \begin{array}{l} (\neg (\Holds(p^1_j,S)\vee \Holds(p^2_j,S))\,\wedge\\ \Holds(turn(i),S)\wedge\neg \Terminal(S))\ \equiv\ \Poss(a^i_j,S) \end{array} \]
\[ \Holds(p^i_j,S)\vee A=a^i_j\equiv \Holds(p^i_j,\Do(A,S)) \]
\[ \begin{array}{ll} \Holds(turn(1),S) \rightarrow & \neg\Holds(turn(1),\Do(A,S))\, \wedge\\ & \Holds(turn(2),\Do(A,S)) \end{array} \]
\[ \begin{array}{ll} \Holds(turn(2),S) \rightarrow & \neg\Holds(turn(2),\Do(A,S))\,\wedge\\ & \Holds(turn(1),\Do(A,S)) \end{array} \]
\end{ContinueExample}

\subsection{Adding Strategy Rules}

As a general logic-based formalism, the Situation Calculus allows for a straightforward encoding of strategy rules with the help of a direct encoding of their interpretation according to~(\ref{eq:sofr}) and Definition~\ref{de:strategyInterpretation}, respectively. 
Based on the rewriting rules from above, the Situation Calculus encoding $\sitcstr{r}{A}{S}$ for a strategy rule $r$ over language $\mathcal{L}$ is inductively obtained as follows, where $A$ and $S$ are variables.
  \[
  \begin{array}{l} \sitcstr{\varphi}{A}{S}\ :=\ \Poss(A,S)\wedge\sitc{\varphi}{A}{S} \\ \\

    \sitcstr{(r_1\ \triangledown\ r_2\,\triangledown \ldots\triangledown\ r_n)}{A}{S}\ := \\
    \ \ \ \ \ \ \sitcstr{r_1}{A}{S} \\
    \ \ \ \ \ \ \vee \\
    \ \ \ \ \ \ \sitcstr{r_2}{A}{S}\,\wedge\,\neg(\exists A')\,\sitcstr{r_1}{A'}{S} \\
    \ \ \ \ \ \ \vee\ \ldots\ \vee \\
    \ \ \ \ \ \ \sitcstr{r_n}{A}{S}\wedge\,\neg(\exists A')\,(\sitcstr{r_1}{A'}{S}\vee\ldots\vee\sitcstr{r_{n-1}}{A'}{S}) \\ \\
    \sitcstr{(r_1\vartriangle r_2\vartriangle \ldots\vartriangle r_n)}{A}{S}\ := \\
    \ \ \ \ \ \ \sitcstr{r_1}{A}{S}\,\wedge\,\neg(\exists A')\,(\sitcstr{r_1}{A'}{S}\wedge \sitcstr{r_2}{A'}{S}) \\
    \ \ \ \ \ \ \vee \\
    \ \ \ \ \ \ \sitcstr{r_1}{A}{S}\wedge\sitcstr{r_2}{A}{S}\,\wedge\,\neg(\exists A')\,(\sitcstr{r_1}{A'}{S}\wedge \sitcstr{r_2}{A'}{S}\wedge \sitcstr{r_3}{A'}{S}) \\
    \ \ \ \ \ \ \vee\ \ldots\ \vee \\
    \ \ \ \ \ \ \sitcstr{r_1}{A}{S}\wedge\sitcstr{r_2}{A}{S}\wedge\ldots\wedge\sitcstr{r_n}{A}{S}
  \end{array}
  \]
%This encoding nicely reflects the duality of the two operators $\triangledown$ and $\vartriangle$, with the former being a form of prioritised disjunction (where the disjunct with higher priority must be satisfied if that is possible) and the latter a form of prioritised conjunction (where the conjunct with lower priority need not be satisfied if that is not possible).

\subsection{Computing with Strategies: Examples}

The representation of strategy rules in the Situation Calculus can be used to define a special predicate, which we denote by $\Strat(\alpha,\sigma)$, whose intended meaning is that it is possible according to some player's strategy to take action~$\alpha$ in situation~$\sigma$. Consider, for example, a given set of complete strategy rules $\{r_1,\ldots,r_n\}$ for each player, then this can be embedded into a Situation Calculus encoding of a game as follows:
\[
  \Strat(A,S)\,\equiv\,\sitcstr{r_1}{A}{S}\vee\ldots\vee\sitcstr{r_n}{A}{S}
\]

%Michael: Why a disjunction of complete strategy rules is needed?

Such information about strategies can be used for a variety of purposes. Specifically, as we will briefly illustrate next, it can be used to infer possible outcomes of a game under a given set of strategy rules or help a player to decide on a course of action by reasoning about opponents' strategies.

\paragraph*{Inferring possible outcomes.}

Based on the predicate $\Strat$, the set of all possible playouts of a game according to players' strategies can be recursively defined as follows.
\[ \StratPlayout(s_0) \]
\[  \StratPlayout(S)\wedge\Strat(A,S)\,\rightarrow\,\StratPlayout(\Do(A,S)) \]
This predicate determines all paths that are reachable if all players follow their given strategy rules. Hence, all possible playouts under these strategies can be determined as all situations~$S$ that satisfy $\StratPlayout(S)\wedge\Terminal(S)$.

\paragraph*{Reasoning about opponents' strategies.}

Another way of using reasoning about strategies is for players to use knowledge or belief about their opponents' strategies in order to compute their own best course of actions in response. As an example, we will consider the encoding of a generalised form of Minimax evaluation in the Situation Calculus. To this end, let us take the perspective of particular player~$i$ and assume that this player's belief about the opponents' strategy rules is encoded using predicate $\Strat$ as above. Let us assume also that $\Turn(i,S)$ expresses the fact that it is player~$i$'s turn in situation~$S$. We can then define recursively the notion of a winning situation for~$i$, represented by predicate $\Winnable(S)$, as follows\/:

\begin{equation}\label{eq:winnable1} \Wins(i,S)\,\rightarrow\,\Winnable(S) \end{equation}
\begin{equation}\label{eq:winnable2} \begin{array}{c} \Turn(i,S)\,\wedge\,(\exists A)(\Poss(A,S)\wedge\Winnable(\Do(A,S))) \\ \rightarrow\,\Winnable(S) \end{array} \end{equation}
\begin{equation}\label{eq:winnable3} \begin{array}{c} \neg\Turn(i,S)\wedge(\forall A)(\Strat(A,S)\rightarrow\Winnable(\Do(A,S)))\\ \rightarrow\,\Winnable(S) \end{array} \end{equation}

According to this definition, a situation is winnable for our player~$i$ if he can choose a course of action whenever it is his turn, (\ref{eq:winnable2}), such that if all other players choose their actions according to their strategy, (\ref{eq:winnable3}), then a terminal situation will be reached in which player~$i$ has won, (\ref{eq:winnable1}).

\subsection{Computing with Strategies in ASP}

\begin{figure}[t]
\ \ \
\begin{minipage}{3.5in}
\begin{lstlisting}
action(a(I,J)).

holds(turn(1),0).
legal(a(I,J),T) :- not holds(p(1,J),T), holds(turn(I),T),
                  not holds(p(2,J),T), not terminal(T).
holds(p(I,J), T+1) :- holds(p(I,J),T).
holds(p(I,J), T+1) :- does(a(I,J)).
holds(turn(1),T+1) :- holds(turn(2),T).
holds(turn(2),T+1) :- holds(turn(1),T).

wins(I,T)   :- holds(p(I,J),T), ..., holds(p(I,J+k),T).
terminal(T) :- wins(I,T).
terminal(T) :- not freecell(T).
freecell(T) :- not holds(p(1,J)), not holds(p(2,J)).

1 { does(A,T) : action(A) } 1.
:- does(A,T), not legal(A,T).

:- non_strategic(T).
non_strategic(T) :- does(A,T), not strat(A,T).

strat(A,T) :- fill_next(A,T).
strat(A,T) :- not exists_fill_next(T), fill_any(A,T).
fill_next(a(I,J),T) :- holds(p(I,J-1),T), legal(a(I,J),T),
                      not holds(p(1,J),T), not holds(p(2,J),T).
fill_next(a(I,J),T) :- holds(p(I,J+1),T), legal(a(I,J),T),
                      not holds(p(1,J),T), not holds(p(2,J),T).
exists_fill_next(T) :- fill_next(A,T).
fill_any(a(I,J),T)  :- legal(a(I,J),T).
\end{lstlisting}
\end{minipage}
\caption{\label{fi:asp} An ASP encoding of the game from Example~\protect\ref{ex:boxdot1}, including a strategy rule. For the sake of brevity, we have omitted the domain definitions for variables $I$, $J$, and $T$ (the latter ranging from 0 to a given time horizon).}
\end{figure}

Under specific conditions, problems that require reasoning about games and strategies can be solved by Answer Set Programming (ASP). This general technique provides a way of computing models for logic programs for which particularly efficient implementations have been developed in the recent past, such as \cite{gebser:potass} just to mention one. ASP has been used successfully for reasoning about actions and plan generation (see, e.g., \cite{lifsch:answer}) as well as for endgame search in general game playing~\cite{B:50}. In this section, we build on these existing methods and show how ASP can be used to compute all possible outcomes of a game under given strategies for the players. We assume the reader to be familiar with basic notions and notations of ASP, as can be found in~\cite{gelfon:answer}.

The standard use of ASP for computing with actions is to replace the branching time structure of the Situation Calculus by linear time.
For deterministic games with complete specification of the initial state and a given time horizon, the game rules can be encoded as an ASP in such a way that each answer set corresponds to a reachable path and vice versa (see, e.g., \cite{B:50}). For our running example game, Figure~\ref{fi:asp} (lines~1--18) constitute an ASP encoding that follows this principle. Specifically, the so-called {\em weight atom\/} in clause~16 requires each answer set to include one, and only one, action at each point in time. The so-called {\em constraint\/} in clause~17 rules out any answer set in which the chosen action is not legal.

A given ASP that provides a linear-time encoding of a game can be extended by encodings of strategy rules for the players so that the answer sets comply with the strategies. This provides a computational method for inferring the possible outcomes of a game under a given set of strategies.

Provided it does not include the \mbox{$\bigcirc$-operator}, any strategy $\varphi$ in our language that can be encoded in this linear way according to the following inductive coding scheme. Let ${\tt n}$ be a unique predicate name that stands for (the satisfaction of)~$\varphi$, then\/:
\[ \begin{array}{c|l}
 \varphi & \multicolumn{1}{c}{\mbox{encoding}} \\ \hline \hline
 p & {\tt n(A,T)}\ \mbox{\tt :-}\ {\tt holds(p, T).} \\
 \neg\psi_1 & {\tt n(A,T)}\ \mbox{\tt :-}\ {\tt not\ n_1(A, T).} \\
 \psi_1\wedge\psi_2 & {\tt n(A,T)}\ \mbox{\tt :-}\ {\tt n_1(A, T),\ n_2(A, T).} \\
 does(a) & {\tt n(A,T)}\ \mbox{\tt :-}\ {\tt A=}\,a{\tt.} \\
 legal(a) & {\tt n(A,T)}\ \mbox{\tt :-}\ {\tt legal(}a{\tt,T).} \\
 wins(i) & {\tt n(A,T)}\ \mbox{\tt :-}\ {\tt wins(}i{\tt,T).} \\
 init & {\tt n(A,T)}\ \mbox{\tt :-}\ {\tt T=0.} \\
 terminal & {\tt n(A,T)}\ \mbox{\tt :-}\ {\tt terminal(T).} \\ \hline
\psi_1\,\triangledown\,\psi_2\,\triangledown\ldots\,\triangledown\,\psi_m
 & {\tt n(A,T)}\ \mbox{\tt :-}\ {\tt n_1(A, T).} \\
 & {\tt n(A,T)}\ \mbox{\tt :-}\ {\tt n_2(A, T),\ not\ {n'}(T).} \\
 & {\tt n(A,T)}\ \mbox{\tt :-}\ {\tt n_3(A, T),\ not\ {n''}(T).} \\
 & \ldots \\
 & {\tt {n'}(T)}\,\ \mbox{\tt :-}\ {\tt {n}_1(A',T).} \\
 & {\tt {n''}(T)}\ \mbox{\tt :-}\ {\tt {n}_2(A',T).} \\
 & \ldots \\ \hline
 \psi_1\vartriangle\psi_2\vartriangle\ldots\vartriangle\psi_m
 & {\tt n(A,T)}\ \mbox{\tt :-}\ {\tt n_1(A, T),\ not\ {n'}(T).} \\
 & {\tt n(A,T)}\ \mbox{\tt :-}\ {\tt n_1(A, T),\ n_2(A, T),\ not\ {n''}(T).} \\
 & \ldots \\
 & {\tt {n'}(T)}\,\ \mbox{\tt :-}\ {\tt {n}_1(A',T),\ {n}_2(A',T).} \\
 & {\tt {n''}(T)}\ \mbox{\tt :-}\ {\tt {n}_1(A',T),\ {n}_2(A',T),\ {n}_3(A',T).} \\
 & \ldots
\end{array} \]
Where necessary, this is accompanied by clauses with head $\tt n_i$ to encode the sub-formulas $\psi_i$ (for $i=1,2,\ldots$), which are obtained inductively.

Clauses~19--29 in Figure~\ref{fi:asp} are an example of applying this encoding in order to constrain the answer sets for our {\it CrossDot} game to those where both players~$i$ follow the simple strategy $(fill\_next^i)\triangledown(\fillany^i)$ (cf.\ Example~\ref{ex:boxdot5}): Constraint~19 in conjunction with clause~20 rejects all answer sets that do not comply with the strategy definition. Clauses~24--27 encode the first part of the strategy rule for both players (in a slightly more compact form than obtained by strictly applying the coding scheme from above), and clause~23, in conjunction with clauses~28--29, encodes the second part.

\subsection{Further extensions and computational complexity}
The encoding scheme from above does not extend to strategies that include the $\bigcirc$-operator since their evaluation requires a counterfactual look\-ahead. Hence, they cannot be directly represented in an ASP based on a linear time structure. There are two conceivable ways to overcome this limitation.
\begin{enumerate}
  \item If the axiomatisation of a game supports the definition of a regression operator similar to the one in the standard Situation Calculus~\cite{reiter:frame}, then any strategy of the form $\bigcirc\varphi$ can be regressed to a formula~$\varphi'$ that is logically equivalent under the game axioms and contains one less occurrence of the $\bigcirc$-operator. The repeated application of regression will yield a $\bigcirc$-free formula, which then can be encoded in the same way as above.
  \item Alternatively, we can extend the linear time structure to allow for more than one sequence of actions in a single same answer set. A similar approach has been shown to be practically viable for proving epistemic properties in general games using ASPs~\cite{B:68}.
\end{enumerate}
A detailed formalisation and analysis of either solution goes beyond the scope of this paper and is left for future work.\\

Up to now we have demonstrated with two example implementations how our formalism supports automated reasoning. With regard to the efficiency of the implementations, we can consider different aspects regarding both the description of strategies and the problem of reasoning about them.

The complexity of translating a strategy in our language to situation calculus is linear, and so is the translation to ASP for $\bigcirc$-free strategy rules. %Encoding a strategy rule with $\bigcirc$ in an ASP program would require counterfactual lookahead therefore its complexity can be non-linear.
However, generally speaking it is unrealistic to expect high efficiency for a generic strategy reasoning mechanism because in theory, as an extension of GDL, our language can describe any finite game with perfect information. Verifying an arbitrary strategy such as, ``try any possible action to win,'' is equivalent to solving a game and therefore equally complex. It is well known that the complexity of finding a winning strategy for complex games like Japanese Go, which can be specified in GDL, is EXPTIME-complete~\cite{Robson83}.  

Moreover, the computational complexity of solving a game sometimes is independent of the length of the game description and the strategies that are used to solve the game. 
%\footnote{For instance, the complexity for solving the $13\times 13$ Go game is totally different from solving the $19\times 19$ Go game although their description can be exactly the same.} 
Therefore, a complexity analysis for domain-independent strategy reasoning mechanisms can be meaningless.\footnote{For instance, the description of game rules for Chinese Go is quite similar to the one for Japanese Go but their complexity is significantly different.} However, it is possible that certain restrictions on both game descriptions and strategy rules may lead to specific upper bounds for the computational complexity of solving these games. Restrictions on the number of lookahead steps can also affect the complexity of reasoning. In addition, since our strategies represent some ideas of how to play a game well, it is possible that we express these ideas in our language and design specific algorithms to automatically generate possible moves that correspond to the ideas. The efficiency of these algorithms is crucial to the design of a game player. We leave these issues for future investigation.

%%%%%%%%%%%%%%%%%%%%%%%%%%%%%%%%%%%%%%%%%%%%%%%%%%%%%%%%%%%%%%%%
\section{Related Work}

Modelling and specifying strategies is a fundamental research theme in game theory. Researchers in artificial intelligence have recently joined in the research but mostly focus on modelling of strategic reasoning with the help of logical approaches. A number of logical frameworks have been proposed in the literature for strategy representation and reasoning~\cite{pauly2002modal,AlurHK02,goranko2006complete,Chatterjee2010677,MogaveroMV10,Benthem2013}. Most of the frameworks were built on either {\it Coalition Logic} (CL),  {\it Alternating-Time Temporal Logic} (ATL), or {\it Propositional Dynamic Logic} (PDL). 

Both coalition logic~\cite{pauly2002modal} and alternating-time temporal logic~\cite{AlurHK02} were developed to model strategic abilities of coalitions in multi-agent systems. The modality $\langle C\rangle\varphi$ (or $\langle C\rangle\!\bigcirc\!\varphi$) expresses that ``a group of agents, $C$, has a joint strategy to bring about $\varphi$ no matter what strategies the other agents choose''. In CL, a strategy of a player is simply an action available to the player. In ATL, a strategy is a function that maps a sequence of states to an action (in ATL). In both logics, strategies stay on the semantic level without syntactical representation.

A number of extensions of either CL or ATL aim to bring strategies to the syntactical level. \citeauthor{vanderHoek05} (\citeyear{vanderHoek05}) proposed an extension of ATL, named CATL for {\it Counterfactual ATL}, with a variation to the coalition modality, $C_i(\sigma,\varphi)$, representing the counterfactual statement, ``if agent $i$ had committed to a strategy $\sigma$, then $\varphi$ would hold''. Strategies in that framework can be explicitly represented on the syntactical level using dynamic logic-like modalities, even though program connectives of dynamic logic are not allowed to be used for combining strategies. \citeauthor{Waltheretal07} (\citeyear{Waltheretal07}) refine the work of CATL into an axiomatic logical system with a different semantics. However, strategies are still restricted to primitive forms, which means that the combination of strategies is not supported. Similar restrictions have also been applied in several other ATL- or CL-like logical frameworks for strategic reasoning, such as \cite{Chatterjee2010677,MogaveroMV10,lorini2010dynamic}. 

Another approach to strategy representation and reasoning is to treat a strategy as a program so that PDL-style program connectives can be used to combine strategies~\cite{Benthem2001,ramanujam:dynamic,Benthem2013}. 
van Benthem proposed a logical framework, named Temporal Forcing Logic (TFL), with a modality $[\sigma, i]\varphi$, meaning that ``player $i$ applies strategy $\sigma$, against any play of the others, to force the game to a state in which $\varphi$ holds'', where a strategy can be defined as any PDL program. Similar proposal can also be found in \cite{ramanujam:dynamic}. Such an ``intuitive analogue to strategies'' provides a close approximation to strategy representation; nevertheless, a strategy has essential differences from a program, which requires specific ways of composition and reasoning as we have shown in the previous sections.   

We like to stress that our treatment of strategies is different from all of the abovementioned approaches in the following aspects. Firstly, we can use the same propositional formulas for different purposes. A propositional formula with the standard semantics of propositional modal logic can represent properties of the game state and be used in domain-dependent axioms. But we also represent a strategy with the help of a propositional formula, by endowing the formula with a specific semantics. This makes strategy design much easier and efficient. Secondly, we view a strategy as a set of possible moves, i.e., a set of state-action pairs, rather than a function from a state (or a sequence of states) to an action. In this sense, our strategies represent ``rough ideas'', which can then be combined and refined. Thirdly, instead of using PDL-style program connectives, we introduced two prioritised connectives for combining strategies, which, as we have seen, provides for a very natural and convenient design of strategies.

We want to mention that the idea of the prioritised disjunction was inspired by \citeauthor{BrewkaBB04}'s (\citeyear{BrewkaBB04}) Qualitative Choice Logic (QCL). QCL contains a non-standard propositional connective $A\overrightarrow{\times}B$ with the meaning,  ``{\it $A$ if possible; but if $A$ is impossible then at least $B$}''. We found that the semantics of the connective fits strategies very well. It is an interesting question for future work whether our prioritised conjunction can be integrated into QCL.
 
\section{Conclusion}

In this paper we have introduced a logical language to describe, compose and combine strategies for game-playing agents. The language derives from the general Game Description Language (GDL) and extends it by a single temporal operator $\bigcirc$ and two new prioritised connectives: $\triangledown$ and $\vartriangle$. The basic components of GDL facilitate the representation of initial and terminal conditions, winning criteria and legality of actions (i.e., preconditions). The temporal operator allows us to describe the effects of actions. These form the basic language for describing game- and player-specific strategies. The newly introduced connectives allow us to combine simple strategies into more complicated and refined ones.  When we use the language to describe a strategy, we endow it with a specific semantics so that we can compose a strategy in a logical way and the actual moves the strategy represents can be generated in an automatic way. 

We have thoroughly analysed the properties of the language. In fact, the nice properties in particular of the new connectives give us great freedom in practice for strategy design: we can start with a strategy that formalises one specific idea. If it is too restricted, we can extend it with more generic ones using the prioritised disjunction, and if it is too generic, we can refine it with more specific strategies using the prioritised conjunction. We have shown also how strategies can be embedded into existing methods for the design and analysis of intelligent agents in order to solve problems that involve reasoning about strategies, including the computation of possible outcomes of games under given strategies.
\comment{We have designed a simple game to demonstrate how to use the techniques we introduced in the work to compose a strategy with desired properties and how to use our framework to reason about strategies in order to develop a solution to a game.}

Our current implementation for strategy reasoning combines reasoning about actions and change with an encoding of game rules and strategies using Answer Set Programming. It would be more efficient to develop a specific method of model checking based on the structure of our state transition model and the syntax of strategy composition. We leave this for future work.

\acknowledgements
The first author was partially supported by the Australian Research
Council through project DP0988750 and National Natural Science Foundation of China  through projects 61003203 and 61262029. The second author was the recipient of an ARC Future Fellowship (project number~FT\,0991348). He is also affiliated with the University of Western Sydney. We thank the anonymous reviewers for their constructive comments and suggestions. We also thank Guifei Jiang for her valuable input.

%%%%%%%%%%%%%%%%%%%%%%%%%%%%%%%%%%%%%%%%%%%%%%%%%%%%%%%%%%%%%%%%%%%%%%%%%%%%%%%%%%%%%%%%%%%%%%%%%%%%%%%%%%

\bibliographystyle{klunamed}

%\bibliography{../../../E-Documents/DongmoBib}
\end{article}
\end{document}